\documentclass[final]{cvpr}

\usepackage{times}
\usepackage{epsfig}
\usepackage{graphicx}
\usepackage{amsmath}
\usepackage{amssymb}
\usepackage{xcolor}              
\usepackage{verbatim}            
\usepackage{booktabs}
\usepackage{relsize}
\usepackage{multirow}
\usepackage{scrextend}           
\usepackage{array}               
\usepackage{makecell}            
\usepackage{wasysym}             

\allowdisplaybreaks              
\usepackage{subcaption}          
\graphicspath{{images/}}
\usepackage{bm}
\usepackage{cite}                
\usepackage{paralist}            

\usepackage[ruled,linesnumbered]{algorithm2e}  
\SetKwIF{If}{ElseIf}{Else}{if}{}{else if}{else}{}
\SetKwFor{For}{for}{do}{end}
\SetKwComment{Comment}{$\triangleright$\ }{}

\SetCommentSty{mycommfont}

\newcommand{\myReferTable}[1]{Tab.~\ref{#1}}
\newcommand{\myReferFigure}[1]{Fig.~\ref{#1}}
\newcommand{\myReferSection}[1]{Sec.~\ref{#1}}

\newcommand{\methodName}{GrooMeD-NMS}
\newcommand{\methodNameShort}{GrooMeD}
\newcommand{\methodNameFull}{Grouped Mathematically Differentiable Non-Maximal Suppression}

\newcommand{\kinematicImage}{Kinematic (Image)}
\newcommand{\kinematicVideo}{Kinematic (Video)}

\newcommand{\iou}{IoU}
\newcommand{\twoD}{$2$D}
\newcommand{\threeD}{$3$D}
\newcommand{\twoDMath}{2\text{D}}
\newcommand{\threeDMath}{3\text{D}}
\newcommand{\iouTwoD}{\iou$_{2\text{D}}$}
\newcommand{\iouThreeD}{\iou$_{3\text{D}}$}
\newcommand{\lidar}{LiDAR}
\newcommand{\giouTwoD}{g\iouTwoD}
\newcommand{\giouThreeD}{g\iouThreeD}
\newcommand{\vol}{V}

\newcommand{\ap}{AP}
\newcommand{\apMath}{\text{\ap}}
\newcommand{\apthreeD}{\ap$_{3\text{D}}$}
\newcommand{\apThreeDForty}{\ap$_{3\text{D}|R_{40}}$}
\newcommand{\apBevForty}{\ap$_{\text{BEV}|R_{40}}$}
\newcommand{\apTwoD}{\ap$_{2\text{D}}$}
\newcommand{\apTwoDForty}{\ap$_{2\text{D}|R_{40}}$}
\newcommand{\aploss}{\ap-Loss}
\newcommand{\imageWise}{Imagewise}

\newcommand{\loss}{\mathcal{L}}
\newcommand{\lossBefore}{\loss_{before}}
\newcommand{\lossAfter}{\loss_{after}}
\newcommand{\lossWeigh}{\lambda}

\newcommand{\groundTruthj}{g_l}
\newcommand{\boxes}{\mathcal{B}}
\newcommand{\boxIndex}{i}
\newcommand{\boxMember}{b}
\newcommand{\boxi}{\boxMember_\boxIndex}
\newcommand{\boxj}{\boxMember_j}
\newcommand{\numboxes}{n}
\newcommand{\boxesAfterNMS}{\mathcal{D}}
\newcommand{\vindex}{\mathbf{d}}

\newcommand{\boxConfidence}{b_{conf}}
\newcommand{\class}{{class}}

\newcommand{\nmsThresh}{N_t}
\newcommand{\validBoxThresh}{v}
\newcommand{\prune}{p}
\newcommand{\basic}{Linear}
\newcommand{\exponentialPruning}{Exponential}
\newcommand{\pruneof}[1]{\prune(#1)}
\newcommand{\temperature}{\tau}
\newcommand{\clip}[1]{\left\lfloor#1\right\rceil}
\newcommand{\numberElem}[1]{\left|#1\right|}
\newcommand{\elementMul}{\!\odot} 

\newcommand{\rescore}{{\bf r}}
\newcommand{\score}{{\bf s}}
\newcommand{\rescoreMember}{r_i}

\newcommand{\tempIndex}{\mathbf{t}}
\newcommand{\tempIndexTwo}{\mathbf{u}}
\newcommand{\tempIndexThree}{\mathbf{v}}
\newcommand{\tempIndexFour}{\mathbf{w}}
\newcommand{\pruneMat}{{\bf{P}}}
\newcommand{\overlapMat}{\mathbf{O}}
\newcommand{\zeroVector}{\mathbf{0}}

\def\lowertriangle{
\resizebox{.015\textwidth}{!}{
\setlength{\unitlength}{.1cm}
\begin{picture}(7,5)(0,0)
\put(6,0){\line(-1,0){6}}
\put(0,0){\line(0,1){5}}
\put(6,0){\line(-6,5){6}}
\end{picture}
}
}
\newcommand{\ourTriangle}{\triangle}

\newcommand{\overlapMatLower}{\overlapMat_{\lowertriangle}}
\newcommand{\overlap}{o}
\newcommand{\identity}{{\bf{I}}}
\newcommand{\mask}{{\bf{M}}}

\newcommand{\groups}{\mathcal{G}}
\newcommand{\groupIndex}{k}
\newcommand{\groupMember}{\groups_\groupIndex}
\newcommand{\boxesGroup}{\boxes_{\groupMember}}
\newcommand{\boxesGroupTwo}{\boxes_{\groups_l}}
\newcommand{\rescoreGroup}{\rescore_{\groupMember}}
\newcommand{\scoreGroup}{\score_{\groupMember}}
\newcommand{\pruneMatGroup}{\pruneMat_{\groupMember}}
\newcommand{\identityGroup}{\identity_{\groupMember}}
\newcommand{\maskGroup}{\mask_{\groupMember}}

\newcommand{\groupSize}{\alpha}

\newcommand{\classicalNmsShort}{classical}
\newcommand{\classicalNmsShortCaps}{Classical}
\newcommand{\classicalNms}{classical NMS}
\newcommand{\classicalNmsCaps}{Classical NMS}
\newcommand{\softNmsCaps}{Soft-NMS}
\newcommand{\softNmsShortCaps}{Soft}
\newcommand{\distanceNmsCaps}{Distance-NMS}
\newcommand{\distanceNmsShortCaps}{Distance}

\newcommand{\bigO}[1]{\mathcal{O}\left(#1\right)}

\newcommand{\thatIs}{\textit{i.e.}}

\newcommand{\argmax}{\operatornamewithlimits{argmax}}

\newcommand{\kitti}{KITTI}
\newcommand{\valOne}{Val 1}
\newcommand{\valTwo}{Val 2}

\newcommand{\myTopRule}{\Xhline{2\arrayrulewidth}}

\newcolumntype{t}{!{\vrule width 1.5\arrayrulewidth}}
\newcolumntype{m}{!{\vrule width 2.5\arrayrulewidth}}

\providecommand\rightarrowRHD{\relbar\joinrel\mathrel\RHD}

\providecommand\longrightarrowRHD{\relbar\joinrel\relbar\joinrel\mathrel\RHD}

\newcommand{\first}[1]{$\textcolor{red}{\mathbf{#1}}$}
\newcommand{\second}[1]{$\textcolor{blue}{\mathbf{#1}}$}
\newcommand{\firstkey}[1]{\textcolor{red}{\textbf{#1}}}
\newcommand{\secondkey}[1]{\textcolor{blue}{\textbf{#1}}}
\newcommand{\sota}{SoTA}
\newcommand{\best}[1]{\mathbf{#1}}
\newcommand{\bestKey}[1]{\textbf{#1}}

\usepackage{pifont}
\newcommand{\cmark}{\checkmark}
\newcommand{\xmark}{\ding{53}}%

\newcommand{\autoBraces}[1]{\left(#1\right)}
\definecolor{XLcolor}{rgb}{0.858, 0.188, 0.478}

\makeatletter
\@namedef{ver@everyshi.sty}{}
\makeatother
\usepackage{tikz} 
\usetikzlibrary{calc}  
\usetikzlibrary{arrows}
\usetikzlibrary{arrows.meta}
\usetikzlibrary{shapes.arrows}
\usetikzlibrary{fadings, shadows}
\usetikzlibrary{decorations.text}
\definecolor{my_green}{rgb}{0.0, 0.9, 0.24}
\definecolor{my_green_2}{rgb}{0.0, 0.4, 0.0}
\definecolor{my_violet}{rgb}{0.79, 0.40, 1} 
\definecolor{my_blue}{rgb}{0.2, 0.6, 1}
\definecolor{my_yellow}{rgb}{0.9, 0.8, 0.54}
\definecolor{my_yellow_2}{rgb}{1, 0.75, 0.}
\definecolor{my_red}{rgb}{1,0,0}
\definecolor{my_purple}{rgb}{0.27,0.8, 0.8}
\definecolor{new_green}{rgb}{0.75,0.97,0.44}
\definecolor{my_orange}{rgb}{1.0,0.6,0.35}
\definecolor{my_orange_2}{rgb}{1.0, 0.2, 0.6} 
\definecolor{backward_color}{rgb}{1.0, 0.6, 0.2}  
\definecolor{forward_color}{rgb}{0.2, 1.0, 0.6}
\def\centerarc[#1](#2)(#3:#4:#5)
    { \draw[#1] ($(#2)+({#5*cos(#3)},{#5*sin(#3)})$) arc (#3:#4:#5); }

\usepackage[pagebackref=true,breaklinks=true,colorlinks,bookmarks=false]{hyperref}

\def\cvprPaperID{5545} 
\def\confYear{CVPR 2021}

\begin{document}

\title{\methodName: Grouped Mathematically Differentiable NMS for Monocular 3D Object Detection}
\author{Abhinav Kumar, Garrick Brazil, Xiaoming Liu\\
Michigan State University, East Lansing, MI, USA\\
{\tt\small $[$kumarab6, brazilga, liuxm$]$@msu.edu }\\
{\small \url{https://github.com/abhi1kumar/groomed_nms}}
}

\twocolumn[{%
\renewcommand\twocolumn[1][]{#1}%
\maketitle
\thispagestyle{empty}
\vspace{-1mm}
\noindent\begin{minipage}{\linewidth} 
    \centering
    \begin{minipage}[t]{.33\linewidth}
        \vspace{0cm}
        \begin{tikzpicture}[scale=0.28, every node/.style={scale=0.50}, every edge/.style={scale=0.50}]
\tikzset{vertex/.style = {shape=circle, draw=black!70, line width=0.06em, minimum size=1.4em}}
\tikzset{edge/.style = {-{Triangle[angle=60:.06cm 1]},> = latex'}}

    \node[inner sep=0pt] (input) at (1.2,0) {\includegraphics[trim={20.9cm 4.2cm 18.75cm 5.6cm},clip,height=1.1cm]{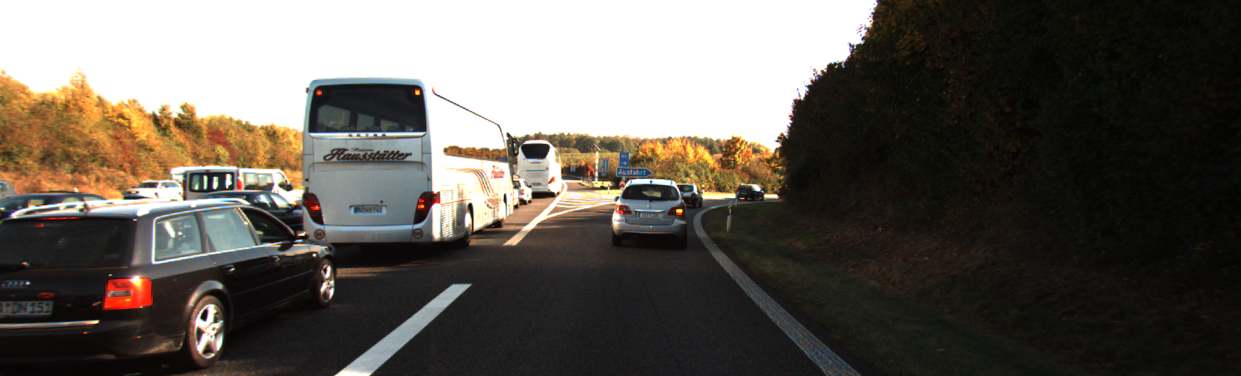}};
    
    \draw [draw=black, fill= my_yellow!40, line width=0.06em, thick] 
       (2.5, 3.0) node[]{}
    -- (2.5,-3.0) node[pos= 0.8, right, scale= 1.25, rotate=90, align=center]{~\\~\\Backbone}
    -- (4,-2.0) node[]{}
    -- (4, 2.0) node[]{}
    -- cycle;
    
    \draw [draw=black, fill= my_blue!40, line width=0.06em, thick] (5, 1.8) rectangle (7.5,0.8) node[pos=0.5, scale= 1.25]{Score};
    \draw [-{Triangle[angle=60:.1cm 1]}, draw=my_blue!50, line width=0.1em, shorten <=0.5pt, shorten >=0.5pt, >=stealth]
       (4.0,1.2) node[]{}
    -- (5.0,1.2) node[]{};

    \draw [draw=black, fill= blue!20, line width=0.06em, thick] (5, 0.5) rectangle (7.5, -0.5) node[pos=0.5, scale= 1.25]{$\twoD$};
    \draw [-{Triangle[angle=60:.1cm 1]}, draw=blue!30, line width=0.1em, shorten <=0.5pt, shorten >=0.5pt, >=stealth]
       (4.0,0) node[]{}
    -- (5.0,0) node[]{};
    
    \draw [draw=black, fill= my_violet!20, line width=0.06em, thick] (5, -0.8) rectangle (7.5, -1.8) node[pos=0.5, scale= 1.25]{$\threeD$};
    \draw [-{Triangle[angle=60:.1cm 1]}, draw=my_violet!40, line width=0.1em, shorten <=0.5pt, shorten >=0.5pt, >=stealth]
       (4.0,-1.3) node[]{}
    -- (5.0,-1.3) node[]{};

    \draw [-{Triangle[angle=60:.1cm 1]}, draw=my_blue!50, line width=0.1em, shorten <=0.5pt, shorten >=0.5pt, >=stealth]
       (7.5,1.2) node[]{}
    -- (12.25,1.2) node[pos=0.84, scale= 1.25, align=center]{$\score$\\};
    \draw [-{Triangle[angle=60:.1cm 1]}, draw=blue!30, line width=0.1em, shorten <=0.5pt, shorten >=0.5pt, >=stealth]
       (7.5,0) node[]{}
    -- (12.25 ,0) node[pos=0.84, scale= 1.25, align=center]{$\overlapMat$\\};
    \draw [, draw=my_violet!40, line width=0.1em, shorten <=0.5pt, shorten >=0.5pt, >=stealth]
       (7.5,-1.2) node[]{}
    -- (8.1,-1.2) node[]{};

    \draw [-{Triangle[angle=60:.1cm 1]}, draw=my_blue!50, line width=0.1em, shorten <=0.5pt, shorten >=0.5pt, >=stealth]
       (9.0,1.2) node[]{}
    -- (9.0,-2.8) node[]{};
    \draw [-{Triangle[angle=60:.1cm 1]}, draw=blue!30, line width=0.1em, shorten <=0.5pt, shorten >=0.5pt, >=stealth]
       (8.5,0) node[]{}
    -- (8.5,-2.7) node[]{};
    \draw [-{Triangle[angle=60:.1cm 1]}, draw=my_violet!40, line width=0.1em, shorten <=0.5pt, shorten >=0.5pt, >=stealth]
       (8, -1.2) node[]{}
    -- (8,-2.8) node[]{};
    
    \draw[draw=my_blue!50, fill=my_blue!50, thick](9,1.2) circle (0.15) node[scale= 1.25]{};
    \draw[draw=blue!30, fill=blue!30, thick](8.5,0) circle (0.15) node[scale= 1.25]{};

    \draw [draw=black, fill= my_orange!80, line width=0.06em] (9.5,0.5) rectangle (11,-0.5) node[pos=0.5, scale= 1.25]{\iou};
    
    \draw[draw=black, fill=my_red!30, thick]( 8.5,-4.2) circle (1.5) node[scale= 1.25]{$\lossBefore$};
    
    \draw [draw=green!80, line width=0.06em]                (5.6, 3.2) rectangle (8.15, 4.85) node[]{};
    \draw [draw=cyan, line width=0.06em]                    (5.4, 2.8) rectangle (7.9   , 5 ) node[]{};
    \draw [draw=orange!80, line width=0.06em, thick]        (5.2, 3.0) rectangle (7.7 , 5.25) node[]{};
    \draw [draw=red, line width=0.06em, very thick]         (5  , 3.5) rectangle (7.5 , 5.5   ) node[]{};

    \draw [draw=purple!80, line width=0.06em, thick]        (8.7, 3.0) rectangle (9.20, 5  ) node[]{};
    \draw [draw=my_yellow_2, line width=0.06em, very thick] (8.5, 3.5) rectangle (9   , 5.25) node[]{};

    \node [inner sep=1pt, scale= 1.25] at (6.8, 2.35)  {$\boxes$};
    
    \draw [-{Triangle[angle=60:.1cm 1]}, dashed, draw=black!40, line width=0.1em, shorten <=0.5pt, shorten >=0.5pt, >=stealth]
       (0.5,7) node[]{}
    -- (20,7) node[pos= 0.9, align=center, scale= 1.25]{Inference\\};
    \draw [draw=black, fill= black!70, line width=0.06em] (12.25,1.8) rectangle (15.75,-0.5) node[pos=0.5, scale= 1.25,align=center]{NMS};

    \draw [draw=red, line width=0.06em, very thick]         (15.5  , 3.5) rectangle (18 , 5.5   ) node[]{};
    \draw [draw=my_yellow_2, line width=0.06em, very thick] (19.0, 3.5) rectangle (19.5   , 5.25) node[]{};

    \draw [-{Triangle[angle=60:.1cm 1]}, draw=black!40, line width=0.15em, shorten <=0.5pt, shorten >=0.5pt, >=stealth]
       (15.75,1.2) node[]{}
    -- (20,1.2) node[pos=0.48, scale= 1.25, align=center]{Predictions\\};
    
    \draw [-{Triangle[angle=60:.1cm 1]}, draw=black!40, line width=0.15em, shorten <=0.5pt, shorten >=0.5pt, >=stealth]
       (15.75,0) node[]{}
    -- (20.1,0) node[pos=0.5, scale= 1.25, align=center]{$\rescore$\\};

    \draw [-{Triangle[angle=60:.1cm 1]}, draw=black!80, line width=0.1em, shorten <=0.5pt, shorten >=0.5pt, >=stealth]
       (0.5,6) node[]{}
    -- (10,6) node[pos= 0.8, align=center, scale= 1.25]{Training\\};
    \draw [draw=black!80, line width=0.15em, shorten <=0.5pt, shorten >=0.5pt, >=stealth]
       (10,6.5) node[]{}
    -- (10,5.5) node[pos= 0.8, align=center, scale= 1.25]{};

    \draw[draw=black, fill=my_red!30, thick]( 8.5,-4.2) circle (1.5) node[scale= 1.25]{$\lossBefore$};

\end{tikzpicture}\\
        \vspace{-0.2cm}
        \captionof*{figure}{(a) Conventional NMS Pipeline}
    \end{minipage}%
    \begin{minipage}[t]{.33\linewidth}
        \vspace{0cm}
        \begin{tikzpicture}[scale=0.28, every node/.style={scale=0.50}, every edge/.style={scale=0.50}]
\tikzset{vertex/.style = {shape=circle, draw=black!70, line width=0.06em, minimum size=1.4em}}
\tikzset{edge/.style = {-{Triangle[angle=60:.06cm 1]},> = latex'}}

    \draw [draw=black, fill= forward_color, line width=0.06em] (12.25,1.8) rectangle (15.75,-0.5) node[pos=0.5, scale= 1.25,align=center]{\methodNameShort\\NMS};
    
    \draw [draw=green!80, line width=0.06em,  densely dashed]         (16.1, 3.2) rectangle (18.65, 4.85) node[]{};
    \draw [draw=cyan, line width=0.06em,  densely dashed]             (15.9, 2.8) rectangle (18.4 , 5   ) node[]{};
    \draw [draw=orange!80, line width=0.06em, thick,  densely dashed] (15.7, 3.0) rectangle (18.2 , 5.25) node[]{};
    \draw [draw=red, line width=0.06em, very thick]                   (15.5, 3.5) rectangle (18.0 , 5.5 ) node[]{};
    
    \draw [draw=purple!80, line width=0.06em, thick,  densely dashed] (19.2, 3.0) rectangle (19.70, 5   ) node[]{};
    \draw [draw=my_yellow_2, line width=0.06em, very thick]           (19.0, 3.5) rectangle (19.5 , 5.25) node[]{};
    
    \draw [-{Triangle[angle=60:.1cm 1]}, draw=forward_color, line width=0.15em, shorten <=0.5pt, shorten >=0.5pt, >=stealth]
       (15.75,1.2) node[]{}
    -- (20,1.2) node[pos=0.48, scale= 1.25, align=center]{Predictions\\};
    \draw [-{Triangle[angle=60:.1cm 1]}, draw=forward_color, line width=0.15em, shorten <=0.5pt, shorten >=0.5pt, >=stealth]
       (15.75,0) node[]{}
    -- (20,0) node[pos=0.5, scale= 1.25, align=center]{$\rescore$\\};
    \draw[draw=forward_color, fill=forward_color, thick](17,0) circle (0.15) node[scale= 1.25]{};
    
    \draw [-{Triangle[angle=60:.1cm 1]}, draw=forward_color, line width=0.15em, shorten <=0.5pt, shorten >=0.5pt, >=stealth]
       (17, 0.0) node[]{}
    -- (17,-2.7) node[]{};

    \draw [-{Triangle[angle=60:.1cm 1]}, draw=black!80, line width=0.15em, shorten <=0.5pt, shorten >=0.5pt, >=stealth]
       (0.5,6) node[]{}
    -- (20,6) node[pos= 0.9, align=center, scale= 1.25]{Training\\};

    \draw[draw=black, fill=my_red!60, thick](17.,-4.2) circle (1.5) node[scale= 1.25]{$\lossAfter$};

\end{tikzpicture}\\
        \vspace{-0.2cm}
        \captionof*{figure}{(b) \methodName~Pipeline\vspace{0.1cm}}
    \end{minipage}%
    \begin{minipage}[t]{.33\linewidth}
        \vspace{0.7cm}
        \hspace{0.2cm}\begin{tikzpicture}[scale=0.21, every node/.style={scale=0.39}, every edge/.style={scale=0.39}]
\tikzset{vertex/.style = {shape=circle, draw=black!70, line width=0.12em, minimum size=1.4em}}
\tikzset{edge/.style = {-{Triangle[angle=60:.06cm 1]},> = latex'}}
    
    \draw [-{Triangle[angle=60:.1cm 1]},draw=forward_color, line width=0.1em, shorten <=0.5pt, shorten >=0.5pt, >=stealth]
       (-1.5,5.0) node[]{}
    -- (0,5.0) node[pos= 0, scale= 1.5,align= center]{$\score$~~~~};
    \draw [,draw=forward_color, line width=0.1em, shorten <=0.5pt, shorten >=0.5pt, >=stealth]
       (-0.15,5.0) node[]{}
    -- (17.3,5.0) node[]{};
    \draw [-{Triangle[angle=60:.1cm 1]},draw=forward_color, line width=0.1em, shorten <=0.5pt, shorten >=0.5pt, >=stealth]
       (17.15,5.0) node[]{}
    -- (17.15,1.5) node[pos= 0.7, scale= 1.5]{$\score$~~~~~};

    \draw [-{Triangle[angle=60:.1cm 1]},draw=backward_color, line width=0.1em, shorten <=0.5pt, shorten >=0.5pt, >=stealth,  densely dashed]
       (17.7,5.4) node[]{}
    -- (-1.5,5.4) node[]{};
    \draw [,draw=backward_color, line width=0.1em, shorten <=0.5pt, shorten >=0.5pt, >=stealth, densely dashed]
       (17.55,5.4) node[]{}
    -- (17.55,1.5) node[]{};

    \draw [-{Triangle[angle=60:.1cm 1]},draw=forward_color, line width=0.1em, shorten <=0.5pt, shorten >=0.5pt, >=stealth]
       (-1.5,-4.80) node[]{}
    -- ( 0,-4.80) node[pos= 0, scale= 1.5]{$\overlapMat$~~~~~};
    \draw [-{Triangle[angle=60:.1cm 1]},draw=forward_color, line width=0.1em, shorten <=0.5pt, shorten >=0.5pt, >=stealth]
       (-0.15,-4.8) node[]{}
    -- (9,-4.8) node[]{};
    \draw[draw=forward_color, fill=forward_color, thick](3.7,-4.8) circle (0.2) node[]{};
    \draw [draw=forward_color, line width=0.1em, shorten <=0.5pt, shorten >=0.5pt, >=stealth]
       (3.7, 3.75) node[]{}
    -- (3.7,-4.8) node[]{};
    \draw [draw=forward_color, line width=0.1em, shorten <=0.5pt, shorten >=0.5pt, >=stealth]
       (3.70, 3.6) node[]{}
    -- (14.8, 3.6) node[]{};
    \draw [-{Triangle[angle=60:.1cm 1]},draw=forward_color, line width=0.1em, shorten <=0.5pt, shorten >=0.5pt, >=stealth]
       (14.65, 3.6) node[]{}
    -- (14.65, 1.5) node[]{};
    
    \draw [-{Triangle[angle=60:.1cm 1]}, draw=backward_color, line width=0.1em, shorten <=0.5pt, shorten >=0.5pt, >=stealth, densely dashed ]
       (3.45,-4.4) node[]{}
    -- (-1.5,-4.4) node[]{};
    \draw [, draw=backward_color, line width=0.1em, shorten <=0.5pt, shorten >=0.5pt, >=stealth, densely dashed]
       (3.3, 4.15) node[]{}
    -- (3.3,-4.4) node[]{};
    \draw [, draw=backward_color, line width=0.1em, shorten <=0.5pt, shorten >=0.5pt, >=stealth, densely dashed]
       (3.3, 4.0) node[]{}
    -- (15.2, 4.0) node[]{};
    \draw [draw=backward_color, line width=0.1em, shorten <=0.5pt, shorten >=0.5pt, >=stealth, densely dashed]
       (15.05, 4.1) node[]{}
    -- (15.05, 1.5) node[]{};

    \draw [draw=black,  fill= yellow!30, line width=0.06em] (0.5, 5.8) rectangle (2.5, -5.8   ) node[pos= 0.5, scale= 1.5, rotate= 00, align=center]{Sort};

    \draw [draw=black, fill=black!40, line width=0.06em, thick] (9, -3.8) rectangle (13, -5.8   ) node[pos=0.5, scale= 1.5]{Group};   
    
    \draw [, draw=black!40, line width=0.1em, shorten <=0.5pt, shorten >=0.5pt, >=stealth,]
       (6.8, -2.8) node[]{}
    -- (17.5, -2.8   ) node[]{};
    \draw [-{Triangle[angle=60:.1cm 1]}, draw=black!40, line width=0.1em, shorten <=0.5pt, shorten >=0.5pt, >=stealth,]
       (6.95, -2.8) node[]{}
    -- (6.95, -1.5) node[]{};
    \draw[draw=black!40, fill=black!40, thick](11.05,-2.8) circle (0.2) node[]{};
    \draw[draw=black!40, fill=black!40, thick](14.3,-2.8) circle (0.2) node[]{};
    \draw [-{Triangle[angle=60:.1cm 1]}, draw=black!40, line width=0.1em, shorten <=0.5pt, shorten >=0.5pt, >=stealth,]
       (11.05, -3.8) node[]{}
    -- (11.05, -1.5) node[pos= 0.23, scale= 1.5]{~~~~~~$\groups$};
    \draw [-{Triangle[angle=60:.1cm 1]}, draw=black!40, line width=0.1em, shorten <=0.5pt, shorten >=0.5pt, >=stealth,]
       (14.3, -2.8) node[]{}
    -- (14.3, -1.5) node[]{};
    \draw [-{Triangle[angle=60:.1cm 1]}, draw=black!40, line width=0.1em, shorten <=0.5pt, shorten >=0.5pt, >=stealth,]
       (17.35, -2.8) node[]{}
    -- (17.35, -1.5) node[]{};
    
    \draw [draw=black, fill=white, line width=0.06em] (2.7, -0.5) rectangle (4.7, -1.7   ) node[pos= 0.5, scale= 1.2, rotate= 00, align=center]{lower};
    \draw [draw=black, fill=white, line width=0.06em] (2.7, 0.5) rectangle (4.7, 1.7   ) node[pos= 0.5, scale= 1.2, rotate= 00, align=center]{$\prune$};
    
    \node[inner sep=0pt] (input) at (5.25,0) {$\left(\vphantom{\includegraphics[height=1.2cm]{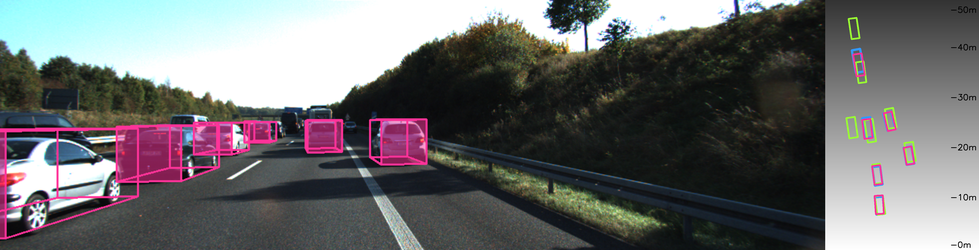}}\right.$};
    
    \draw [draw=black, line width=0.06em] (5.45, 1.5) rectangle (8.45, -1.5   ) node[pos= 0.5, scale= 1.25, rotate= 00, align=center]{};
    \node [inner sep=1pt, scale= 1.5] at (6.8, 2.55)  {$\identity$};
    
    \draw [draw=black, fill=my_blue!20, line width=0.0em] (5.48, 1.47) rectangle (7.45, -0.5   ) node[]{};
    \draw [draw=black, fill=my_orange_2!20, line width=0.0em] (7.45,-0.5) rectangle (8.42, -1.47   ) node[]{};
    
    \draw [draw=black, fill=my_blue, line width=0.0em] (5.48, 1.47) rectangle (5.95, 1.   ) node[]{};    
    \draw [draw=black, fill=my_blue, line width=0.0em] (5.95, 1.) rectangle (6.45, 0.5   ) node[]{}; 
    \draw [draw=black, fill=my_blue, line width=0.0em] (6.45, 0.5) rectangle (6.95, 0.   ) node[]{}; 
    \draw [draw=black, fill=my_blue, line width=0.0em] (6.95, 0.) rectangle (7.45, -0.5   ) node[]{}; 
    \draw [draw=black, fill=my_orange_2, line width=0.0em] (7.45, -0.5) rectangle (7.95, -1.0   ) node[]{}; 
    \draw [draw=black, fill=my_orange_2, line width=0.0em] (7.95, -1.0) rectangle (8.42, -1.47) node[]{}; 
    
    \draw [,draw=black, line width=0.075em, shorten <=0.5pt, shorten >=0.5pt, >=stealth]
       (8.6,0.0) node[]{}
    -- (9.3,0.0) node[]{};

    \draw [draw=black, line width=0.06em] (9.55, 1.5) rectangle (12.55, -1.5   ) node[pos= 0.5, scale= 1.25, rotate= 00, align=center]{};
    \node [inner sep=1pt, scale= 1.5] at (10.9, 2.55)  {$\mask$};
    \draw [draw=black, fill=my_blue!20, line width=0.0em] (9.58, 1.47) rectangle (11.55, -0.5   ) node[]{};
    \draw [draw=black, fill=my_orange_2!20, line width=0.0em]  (11.55,-0.5) rectangle (12.52, -1.47  ) node[]{}; \draw [draw=black, fill=my_blue, line width=0.0em] (9.58, 1.47) rectangle (10.25, -0.5   ) node[]{};
    \draw [draw=black, fill=my_orange_2, line width=0.0em]  (11.55,-0.5) rectangle (12.05, -1.47  ) node[]{}; 
    
    \draw[draw=black](12.95,0) circle (0.25) node[scale= 1.25]{};
    \draw[fill=black](12.95,0) circle (0.07) node[scale= 1.25]{};
    
    \draw [draw=black, fill= white, line width=0.06em] (13.35, 1.5) rectangle (16.35, -1.5   ) node[pos= 0.5, scale= 1.25, rotate= 00, align=center]{};
    \node [inner sep=1pt, scale= 1.5] at (14.2, 2.55)  {$\pruneMat$};
    \draw [draw=black, fill=my_blue!20, line width=0.0em] (13.38, 1.47) rectangle (15.35, -0.5   ) node[]{};
    \draw [draw=black, fill=my_orange_2!20, line width=0.0em]  (15.35,-0.5) rectangle (16.32, -1.47  ) node[]{};
    
    \draw [draw=black, fill=my_blue, line width=0.0em] (13.38, 1) rectangle (13.85, 0.5   ) node[]{};   
    \draw [draw=black, fill=my_blue, line width=0.0em] (13.38, 0.5) rectangle (13.85, 0   ) node[]{};
    \draw [draw=black, fill=my_blue, line width=0.0em] (13.85, 0.5) rectangle (14.35, 0.   ) node[]{};   
    \draw [draw=black, fill=my_blue, line width=0.0em] (13.38, 0) rectangle (13.85, -0.5   ) node[]{};
    \draw [draw=black, fill=my_blue, line width=0.0em] (13.85, 0) rectangle (14.35, -0.5   ) node[]{};
    \draw [draw=black, fill=my_blue, line width=0.0em] (14.35, 0) rectangle (14.85, -0.5   ) node[]{};
    \draw [draw=black, fill=my_orange_2, line width=0.0em] (15.38, -1.0) rectangle (15.85, -1.47   ) node[]{};

    \node[inner sep=0pt] (input) at (16.6,0) {$\left.\vphantom{\includegraphics[height=1.2cm]{images/qualitative/000514.png}}\right)$};
    
    \draw [draw=black, fill= white, line width=0.06em] (17.0, 1.5) rectangle (17.7, -1.5   ) node[pos= 0.5, scale= 1.25, rotate= 00, align=center]{};
    \draw [draw=black, fill=my_blue, line width=0.0em] (17.03, 1.47) rectangle (17.67, -0.5   ) node[]{};
    \draw [draw=black, fill=my_orange_2, line width=0.0em]  (17.03, -0.5) rectangle (17.67, -1.47  ) node[]{};
    
    \draw [, draw=forward_color, line width=0.1em, shorten <=0.5pt, shorten >=0.5pt, >=stealth,]
       (17.7,0.2) node[]{}
    -- (20.4,0.2) node[]{};
    \draw [, draw=backward_color, line width=0.1em, shorten <=0.5pt, shorten >=0.5pt, >=stealth, densely dashed]
       (17.7,-0.2) node[]{}
    -- (20.4,-0.2) node[]{};
    \draw [, draw=backward_color, line width=0.1em, shorten <=0.5pt, shorten >=0.5pt, >=stealth, densely dashed]
       (18.3,-0.) node[]{}
    -- (18.3,-4.6) node[]{};
    \draw [-{Triangle[angle=60:.1cm 1]}, draw=backward_color, line width=0.1em, shorten <=0.5pt, shorten >=0.5pt, >=stealth, densely dashed]
       (15.05, -4.55) node[]{}
    -- (15.05, -1.5) node[]{};
    \draw [draw=black, fill=white, line width=0.06em] (18.9, .6) rectangle (20.1, -.6   ) node[pos= 0.5, scale= 1.2, rotate= 00, align=center]{$\clip{.}$};
    
    \draw [draw=black, fill= white, line width=0.06em] (20.4, 1.5) rectangle (21.1, -1.5   ) node[]{};
    \draw [draw=black, fill=my_blue, line width=0.0em] (20.43, 1.47) rectangle (21.07, -0.5   ) node[]{};
    \draw [draw=black, fill=my_orange_2, line width=0.0em]  (20.43, -0.5) rectangle (21.07, -1.47  ) node[]{};
    
    \draw [, draw=forward_color, line width=0.1em, shorten <=0.5pt, shorten >=0.5pt, >=stealth,]
       (20.55,-1.5) node[]{}
    -- (20.55,-4.95) node[]{};
    \draw [-{Triangle[angle=60:.1cm 1]}, draw=forward_color, line width=0.1em, shorten <=0.5pt, shorten >=0.5pt, >=stealth,]
           (20.45,-4.8) node[]{}
        -- (22.75,-4.8) node[pos= 0.9, scale= 1.5]{~~~~~$\rescore$};
    
    \draw[draw=backward_color, fill=backward_color, thick](18.3,-.2) circle (0.18) node[scale= 1.25]{};
    \draw [, draw=backward_color, line width=0.1em, shorten <=0.5pt, shorten >=0.5pt, >=stealth, densely dashed]
           (18.15,-4.4) node[]{}
        -- (14.90,-4.4) node[]{};
    \draw [, draw=backward_color, line width=0.1em, shorten <=0.5pt, shorten >=0.5pt, >=stealth, densely dashed]
           (22.35,-4.4) node[]{}
        -- (20.85,-4.4) node[]{};
    \draw [, draw=backward_color, line width=0.1em, shorten <=0.5pt, shorten >=0.5pt, >=stealth, densely dashed]
           (20.85,-4.55) node[]{}
        -- (20.85,-1.5) node[]{};
    \draw [-{Triangle[angle=60:.1cm 1]}, draw=backward_color, line width=0.1em, shorten <=0.5pt, shorten >=0.5pt, >=stealth,  densely dashed]
           (22.75,-4.4) node[]{}
        -- (21.50,-4.4) node[]{};
    
    \draw [, draw=forward_color, line width=0.1em, shorten <=0.5pt, shorten >=0.5pt, >=stealth,]
       (20.75,1.5) node[]{}
    -- (20.75,5.15) node[]{};
    \draw [draw=black, fill= white, line width=0.06em] (19.8, 3.5) rectangle (21.3, 4.5   ) node[pos= 0.5, scale= 1.5, rotate= 00, align=center]{$\!>\!v$};
    \draw [-{Triangle[angle=60:.1cm 1]}, draw=forward_color, line width=0.1em, shorten <=0.5pt, shorten >=0.5pt, >=stealth,]
           (20.75,5.0) node[]{}
        -- (22.75,5.0) node[pos= 0.9, scale= 1.5]{~~~~~$\vindex$};    
        
    \draw [,draw=forward_color, line width=0.1em, shorten <=0.5pt, shorten >=0.5pt, >=stealth]
       ( 2.5,-7) node[]{}
    -- ( 4.5,-7) node[pos= 1, scale= 1.5]{~~~~~~~~~~~~~~~~~~~Forward};
    \draw [,draw=backward_color, line width=0.1em, shorten <=0.5pt, shorten >=0.5pt, >=stealth, densely dashed]
       ( 12.5,-7) node[]{}
    -- ( 14.5,-7) node[pos= 1, scale= 1.5]{~~~~~~~~~~~~~~~~~~~Backward};
    
    \draw [draw=black, line width=0.06em, thick] (0, 6.4) rectangle (21.5, -6.4   ) node[]{};

\end{tikzpicture}
\\
        \vspace{0.0cm}
        \captionof*{figure}{(c) \methodName~layer\label{fig:layer}\vspace{0.1cm}}
    \end{minipage}
    
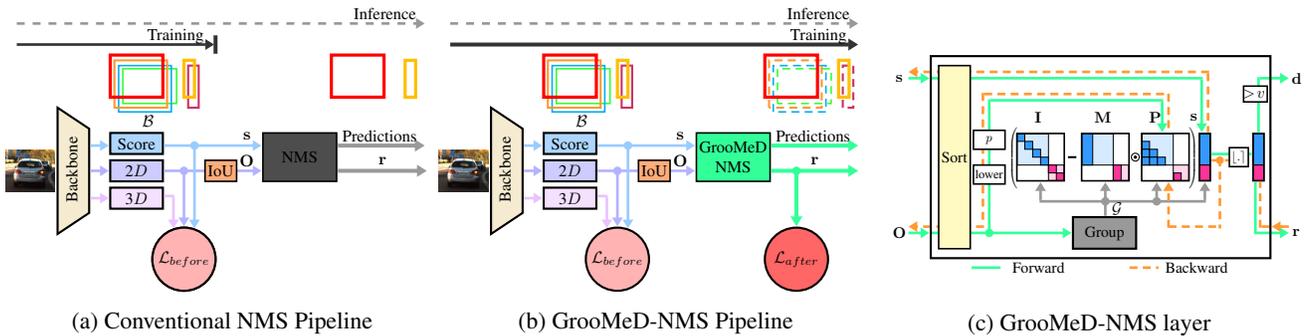
\captionof{figure}{    
    \textbf{Overview of our method.}
    (a) Conventional object detection has a mismatch between training and inference as it uses NMS only in inference. (b) To address this, we propose a novel \methodName~layer, such that the network is trained end-to-end with NMS applied.
    $\score$ and $\rescore$ denote the score of boxes~$\boxes$ before and after the NMS respectively. $\overlapMat$ denotes the matrix containing \iouTwoD~overlaps of $\boxes$. $\lossBefore$ denotes the losses before the NMS,  
    while $\lossAfter$ denotes the loss after the NMS. 
    (c) \methodName~layer calculates~$\rescore$ in a differentiable manner giving gradients from $\lossAfter$ when the best-localized box corresponding to an object is not selected after NMS.
    \vspace{0.4cm}}
    \label{fig:overview}
\end{minipage}
}]

\begin{abstract}
   \vspace{-0.3cm}
   Modern 3D object detectors have immensely benefited from the end-to-end learning idea. 
   However, most of them use a post-processing algorithm called Non-Maximal Suppression (NMS) only during inference. 
   While there were attempts to include NMS in the training pipeline for tasks such as 2D object detection, they have been less widely adopted due to a non-mathematical expression of the NMS. 
   In this paper, we present and integrate GrooMeD-NMS -- a novel 
   Grouped Mathematically Differentiable NMS for monocular 3D object detection, such that the network is trained end-to-end with a loss on the boxes after NMS.
   We first formulate NMS as a matrix operation and then group and mask the boxes in an unsupervised manner to obtain a simple closed-form expression of the NMS. 
   GrooMeD-NMS addresses the mismatch between training and inference pipelines and, therefore, forces the network to select the best 3D box in a differentiable manner. 
   As a result, GrooMeD-NMS achieves state-of-the-art monocular 3D object detection results on the KITTI benchmark dataset performing comparably to monocular video-based methods.
\end{abstract}

\section{Introduction}\label{sec:Introduction}
    \threeD~object detection is one of the fundamental problems in computer vision, where the task is to infer \threeD~information of the object. Its applications include augmented reality~\cite{alhaija2018augmented,rematas2018soccer}, robotics~\cite{saxena2008robotic,levine2018learning}, medical surgery~\cite{rey2002automatic}, and, more recently path planning and scene understanding in autonomous driving~\cite{chen2017multi,huang2020epnet,simonelli2020disentangling,li2020rtm3d}. Most of the \threeD~object detectors 
    ~\cite{chen2017multi,li2019gs3d,huang2020epnet,simonelli2020disentangling,li2020rtm3d} are extensions of the \twoD~object  
    detector Faster R-CNN~\cite{ren2015faster}, which relies on the end-to-end learning idea to achieve State-of-the-Art (\sota)~object detection.
    Some of these methods have proposed changing 
    architectures~\cite{simonelli2020disentangling,shi2020distance,li2020rtm3d} or losses~\cite{brazil2019m3d,chen2020monopair}. 
    Others have tried incorporating confidence~\cite{simonelli2020disentangling,brazil2020kinematic,shi2020distance} 
    or temporal cues~\cite{brazil2020kinematic}.
    
    Almost all of them output a massive number of boxes for each object and, thus, rely on post-processing with a greedy~\cite{prokudin2017learning} clustering algorithm called Non-Maximal Suppression (NMS) during inference to reduce the number of false positives and increase performance. 
    However, these works have largely overlooked NMS's inclusion in training leading to an apparent \emph{mismatch} between training and inference pipelines as the losses are applied on all boxes before NMS but not on \emph{final} boxes after NMS (see \myReferFigure{fig:overview}(a)).\linebreak
    We also find that \threeD~object detection suffers a greater mismatch between classification and \threeD~localization compared to that of \twoD~localization, as discussed further in \myReferSection{sec:results_oracle_additional} of the supplementary and observed in~\cite{huang2020epnet, brazil2020kinematic, shi2020distance}.
    Hence, our focus is \threeD~object detection. 
    
    Earlier attempts to include NMS in the training pipeline~\cite{hosang2016convnet, prokudin2017learning, hosang2017learning} have been made for \twoD~object detection where the improvements are less visible.
    Recent efforts to improve the correlation in \threeD~object detection involve calculating~\cite{simonelli2019disentangling, simonelli2020disentangling} or predicting~\cite{brazil2020kinematic, shi2020distance} the scores via likelihood estimation~\cite{kumar2020luvli} or enforcing the correlation explicitly~\cite{huang2020epnet}. 
    Although this improves the \threeD~detection performance, 
    improvements are limited as their training pipeline is not end to end in the absence of a differentiable NMS. 
    
    To address the mismatch between training and inference pipelines as well as the mismatch between classification and \threeD~localization, we propose including the NMS in the training pipeline, which gives a useful gradient to the network so that it figures out which boxes are the best-localized in \threeD~and, therefore, should be ranked higher (see \myReferFigure{fig:overview}(b)).
    
    An ideal NMS for inclusion in the training pipeline should be not only differentiable but also parallelizable. 
    Unfortunately, the inference-based \classicalNms~and \softNmsCaps~\cite{bodla2017soft} are greedy, set-based and, therefore, not parallelizable~\cite{prokudin2017learning}. 
    To make the NMS parallelizable, we first formulate the \classicalNms~as matrix operation and then obtain a closed-form mathematical expression using elementary matrix operations such as matrix multiplication, matrix inversion, and clipping. 
    We then replace the threshold pruning in the \classicalNms~with its softer version~\cite{bodla2017soft} to get useful gradients. These two changes make the NMS GPU-friendly, and the gradients are backpropagated.  
    We next group and mask the boxes in an unsupervised manner, which removes the matrix inversion and simplifies our proposed differentiable NMS expression further. We call this NMS as~\methodNameFull~(\methodName).

    In summary, the main contributions of this work include:
    \begin{compactitem}
        \item This is the first work to propose and integrate a closed-form mathematically differentiable NMS for object detection, such that the network is trained end-to-end with a loss on the boxes after NMS.
        \item We propose an unsupervised grouping and masking on the boxes to remove the matrix inversion in the closed-form NMS expression.
        \item We achieve \sota~monocular~\threeD~object detection performance on the \kitti~dataset performing comparably to monocular video-based methods.
    \end{compactitem}

\section{Related Work}

    \noindent\textbf{3D~Object Detection.}
        Recent success in \twoD~object detection~\cite{girshick2014rich,girshick2015fast, ren2015faster,redmon2016you,lin2018focal}
        has inspired people to infer \threeD~information from a single \twoD~(monocular) image. 
        However, the monocular problem is ill-posed due to the inherent scale/depth ambiguity~\cite{tang2020center3d}. 
        Hence, approaches use additional sensors such as \lidar~\cite{shi2019pointrcnn,wu2020motionnet,huang2020epnet}, stereo~\cite{li2019stereo,wang2019pseudo} or radar~\cite{vasile2005pose,moosmann2009segmentation}. 
        Although \lidar~depth estimations are accurate, \lidar~data is sparse~\cite{hu2020you} and computationally expensive to process~\cite{tang2020center3d}. 
        Moreover, \lidar s are expensive and do not work well in severe weather~\cite{tang2020center3d}. 
    
        Hence, there have been several works on monocular \threeD~object detection. Earlier approaches~\cite{payet2011contours, fidler20123d, pepik2015multi, chen2016monocular} use hand-crafted features, while the recent ones are all based on deep learning.
        Some of these methods have proposed changing
        architectures~\cite{liu2019deep,li2020rtm3d,tang2020center3d} or losses~\cite{brazil2019m3d,chen2020monopair}. 
        Others have tried incorporating confidence~\cite{liu2019deep,simonelli2020disentangling,brazil2020kinematic,shi2020distance}, augmentation~\cite{simonelli2020towards},  
        depth in convolution~\cite{brazil2019m3d, ding2020learning} or temporal cues~\cite{brazil2020kinematic}. 
        Our work proposes to incorporate NMS in the training pipeline of monocular \threeD~object detection.
    
    \noindent\textbf{Non-Maximal Suppression.}
        NMS has been used to reduce false positives in edge detection~\cite{rosenfeld1971edge}, feature point detection~\cite{harris1988combined, lowe2004distinctive, mikolajczyk2004scale}, face detection~\cite{viola2001rapid}, human detection~\cite{dalal2005histograms, brazil2017illuminating, brazil2019pedestrian} as well as \sota~\twoD~\cite{girshick2015fast, ren2015faster,redmon2016you,lin2018focal} and \threeD~ detection~\cite{bao2019monofenet, chen2017multi, simonelli2020disentangling, brazil2020kinematic, shi2020distance, tang2020center3d}.
        Modifications to NMS in \twoD~detection~\cite{desai2011discriminative, bodla2017soft,hosang2016convnet, prokudin2017learning, hosang2017learning},~\twoD~pedestrian detection~\cite{rujikietgumjorn2013optimized, lee2016individualness, liu2019adaptive},~\twoD~salient object detection~\cite{zhang2016unconstrained} and \threeD~detection~\cite{shi2020distance} can be classified into three categories -- inference NMS~\cite{bodla2017soft,shi2020distance}, optimization-based NMS~\cite{desai2011discriminative,wan2015end, rujikietgumjorn2013optimized, lee2016individualness, zhang2016unconstrained, azadi2017learning} and neural network based NMS~\cite{henderson2016end, hosang2016convnet, prokudin2017learning, hosang2017learning, liu2019adaptive}. 
        
        The inference NMS~\cite{bodla2017soft} changes the way the boxes are pruned in the final set of predictions. 
        \cite{shi2020distance} uses weighted averaging to update the $z$-coordinate after NMS.
        \cite{rujikietgumjorn2013optimized} solves quadratic unconstrained binary optimization while~\cite{lee2016individualness, azadi2017learning, some2020determinantal} and~\cite{zhang2016unconstrained} use point processes and MAP based inference respectively. 
        \cite{desai2011discriminative} and~\cite{wan2015end} formulate NMS as a structured prediction task for isolated and all object instances respectively.
        The neural network NMS use a multi-layer network and message-passing to approximate NMS~\cite{hosang2016convnet, prokudin2017learning, hosang2017learning} or to predict the NMS threshold adaptively~\cite{liu2019adaptive}.
        \cite{henderson2016end} approximates the sub-gradients of the network without modelling NMS via a transitive relationship. 
        Our work proposes a grouped closed-form mathematical approximation of the \classicalNms~and does not require multiple layers or message-passing. 
        We detail these differences in \myReferSection{sec:difference}.

        \begin{algorithm}[!t]
            \small
            \caption{\classicalNmsShortCaps/\softNmsCaps~\cite{bodla2017soft}}
            \label{alg:classical}
            \SetAlgoLined
            \DontPrintSemicolon
            \KwIn{$\score$:~scores, $\overlapMat$:~\iouTwoD~matrix, $N_t$:~NMS threshold,
                    $\prune$: pruning function, $\temperature$:~temperature
                    }
            \KwOut{ $\vindex$: box index after NMS, $\rescore$: scores after NMS}
            \Begin
            {
                $\vindex \gets \{ \}$\;
                $\tempIndex \gets \{1, \cdots, \numberElem{\score}\}$ \Comment*[r]{All box indices}
                $\rescore \gets \score$\;
                \While{$\tempIndex \ne empty$}
                {
                    $\nu \gets \argmax~\rescore[\tempIndex]$ \Comment*[r]{Top scored box}
                    $\vindex \gets \vindex \cup \nu$ \Comment*[r]{Add to valid box index}
                    $\tempIndex \gets \tempIndex - \nu$ \Comment*[r]{Remove from $\tempIndex$}
                    \For{$\boxIndex \gets 1:\numberElem{\tempIndex}$}
                    {
                            $\rescoreMember \gets (1- p_{\temperature}(\overlapMat[\nu, i]))\rescoreMember $ \Comment*[r]{Rescore}
                    }
                }
            }
        \end{algorithm}

\section{Background}

    \subsection{Notations}\label{subsec:notations}
        Let $\boxes\!=\!\{b_i\}_{i=1}^\numboxes$ denote the set of boxes or proposals $\boxi$ from an image.
        Let $\score\!=\!\{s_i\}_{i=1}^\numboxes$ and $\rescore\!=\!\{r_i\}_{i=1}^\numboxes$ denote their scores (before NMS) and rescores (updated scores after NMS) respectively such that $r_i, s_i \ge 0~\forall~i$. $\boxesAfterNMS$ denotes the subset of $\boxes$ after the NMS. 
        Let $\overlapMat= [\overlap_{ij}]$  denote the $\numboxes\times \numboxes$ matrix with $\overlap_{ij}$ denoting the \twoD~Intersection over Union~(\iouTwoD)~of $\boxi$ and $\boxj$. 
        The \emph{pruning} function $\prune$ decides how to rescore a set of boxes $\boxes$ based on \iouTwoD~overlaps of its neighbors, sometimes suppressing boxes entirely.
        In other words, $\pruneof{\overlap_i} = 1$ denotes the box $\boxi$ is suppressed while $\pruneof{\overlap_i} = 0$ denotes $\boxi$ is kept in $\boxesAfterNMS$. 
        The NMS threshold $\nmsThresh$ is the threshold for which two boxes need in order for the non-maximum to be suppressed. 
        The temperature $\temperature$ controls the shape of the exponential and sigmoidal pruning functions $\prune$. 
        $\validBoxThresh$ thresholds the rescores in \methodNameShort~and \softNmsCaps~\cite{bodla2017code} to decide if the box remains valid after NMS.
        
        $\boxes$ is partitioned into different groups $\groups\!=\!\{\groupMember\}$. 
        $\boxesGroup$ denotes the subset of $\boxes$ belonging to group $\groupIndex$. 
        Thus, $\boxesGroup\!=\!\{\boxi\}~\forall~\boxi \in \groupMember$ and $ \boxesGroup\cap\boxesGroupTwo\!=\!\phi~\forall~\groupIndex \ne l$. 
        $\groupMember$ in the subscript of a variable denotes its subset corresponding to $\boxesGroup$. 
        Thus, $\scoreGroup$ and $\rescoreGroup$ denote the scores and the rescores of $\boxesGroup$ respectively. 
        $\alpha$ denotes the maximum group size.
        
        $\vee$ denotes the logical OR while $\clip{x}$ denotes clipping of $x$ in the range $[0,1]$. Formally,
        \vspace{-0.2cm}
        \begin{align}
            \clip{x} &= 
            \begin{cases}
                1, & x > 1\\
                x, & 0 \le x \le 1\\
                0, & x < 0
            \end{cases}
        \vspace{-0.2cm}
        \end{align}
        $\numberElem{\score}$ denotes the number of elements in $\score$. 
        $\lowertriangle$ in the subscript denotes the lower triangular version of the matrix without the principal diagonal.
        $\elementMul$ denotes the element-wise multiplication.
        $\identity$ denotes the identity matrix.

    \subsection{\classicalNmsShortCaps~and~\softNmsCaps}
        NMS is one of the building blocks in object detection whose
        high-level goal is to iteratively suppress boxes which have too much \iou~with a nearby high-scoring box.
        We first give an overview of the classical and \softNmsCaps~\cite{bodla2017soft}, which are greedy and used in inference. 
        \classicalNmsCaps~uses the idea that the score of a box having a high \iouTwoD~overlap with \emph{any} of the selected boxes should be suppressed to zero. 
        That is, it uses a hard pruning $p$ without any temperature $\temperature$. 
        \softNmsCaps~makes this pruning soft via temperature $\temperature$. 
        Thus, \classicalNmsShort~and \softNmsCaps~only differ in the choice of $p$. 
        We reproduce them in Alg.~\ref{alg:classical} using our notations.
        \begin{algorithm}[!t]
            \small
            \caption{\methodName}
            \label{alg:diff_nms}
            \SetAlgoLined
            \DontPrintSemicolon
            \KwIn{$\score$: scores, $\overlapMat$: \iouTwoD~matrix, $\nmsThresh$: NMS threshold,
                    $\prune$: pruning function,
                    $\validBoxThresh$: valid box threshold, $\groupSize$: maximum group size}
            \KwOut{ $\vindex$: box index after NMS, $\rescore$: scores after NMS}
            \Begin
            {
                $\score$, index $\gets \text{sort}(\score$, descending$=$ True$)$\Comment*[r]{Sort $\score$}
                $\overlapMat \gets \overlapMat[$index$][:$, index$]$ \Comment*[r]{Sort $\overlapMat$}
                $\overlapMatLower \gets \text{lower}(\overlapMat)$ \Comment*[r]{Lower $\ourTriangle$ular matrix}
                $\pruneMat \gets \prune(\overlapMatLower)$ \Comment*[r]{Prune matrix}
                $\identity \gets \text{Identity}(\numberElem{\score})$ \Comment*[r]{Identity matrix}
                $\groups \gets \text{group}(\overlapMat, \nmsThresh, \groupSize)$ \Comment*[r]{Group boxes $\boxes$}
                \For{$\groupIndex \gets 1:\numberElem{\groups}$}
                {
                    $\maskGroup \gets$ zeros ($\numberElem{\groupMember}, \numberElem{\groupMember}$)\Comment*[r]{Prepare mask}
                    $\maskGroup[:, \groupMember[1]] \gets 1$ \Comment*[r]{First col of $\maskGroup$}
                    $\rescore_{\groupMember} \gets \clip{\left( \identityGroup - \maskGroup \elementMul \pruneMatGroup \right) \scoreGroup}$ \Comment*[r]{Rescore}
                }
                $\vindex \gets \text{index}[ \rescore >= \validBoxThresh]$ \Comment*[r]{Valid box index}
            }
        \end{algorithm}
        \begin{algorithm}[!t]
            \small
            \caption{Grouping of boxes}
            \label{alg:group}
            \SetAlgoLined
            \DontPrintSemicolon
            \KwIn{$\overlapMat$: sorted \iouTwoD~matrix, $\nmsThresh$: NMS threshold,
                    $\groupSize$: maximum group size}
            \KwOut{ $\groups$: Groups}
            \Begin{
                    $\groups \gets \{ \}$\;
                    $\tempIndex \gets \{1,\cdots,\overlapMat.\text{shape}[1]\}$\Comment*[r]{All box indices}
                    \While{$\tempIndex \ne empty$}
                    {
                        $\tempIndexTwo\!\gets\!\overlapMat[:,1]\!>\nmsThresh$ \Comment*[r]{High overlap indices}
                        $\tempIndexThree \gets \tempIndex[\tempIndexTwo]$\Comment*[r]{New group}
                        $n_{\groupMember} \gets \min(|\tempIndexThree|, \alpha)$\;
                        $\groups$.insert$(\tempIndexThree[:n_{\groupMember}])$\Comment*[r]{Insert new group}
                        $\tempIndexFour\!\gets\!\overlapMat[:,1]\!\le\nmsThresh$\Comment*[r]{Low overlap indices}
                        $\tempIndex \gets \tempIndex[\tempIndexFour]$\Comment*[r]{Keep $\tempIndexFour$ indices in $\tempIndex$}
                        $\overlapMat \gets \overlapMat[\tempIndexFour][:,\tempIndexFour]$\Comment*[r]{Keep $\tempIndexFour$ indices in $\overlapMat$}
                }
            }
        \end{algorithm}

\section{\methodName}
    \classicalNmsCaps~(Alg.~\ref{alg:classical}) uses $\argmax$ and greedily calculates the rescore $r_i$ of boxes $\boxes$ and, is thus not parallelizable or differentiable~\cite{prokudin2017learning}.
    We wish to find its smooth approximation in closed-form for including in the training pipeline.

    \subsection{Formulation}

        \subsubsection{Sorting}
            \classicalNmsCaps~uses the non-differentiable hard $\argmax$ operation (Line $6$ of Alg.~\ref{alg:classical}). 
            We remove the $\argmax$ by hard sorting the scores $\score$ and $\overlapMat$ in decreasing order (lines $2$-$3$ of Alg.~\ref{alg:diff_nms}).
            We also try making the sorting soft. 
            Note that we require the permutation of $\score$ to sort $\overlapMat$. 
            Most soft sorting methods~\cite{ poganvcic2019differentiation,paulus2020gradient,blondel2020fast,berthet2020learning} apply the soft permutation to the same vector. 
            Only two other methods~\cite{cuturi2019differentiable,prillo2020softsort} can apply the soft permutation to another vector. Both methods use $\bigO{n^2}$ computations for soft sorting~\cite{blondel2020fast}. 
            We implement~\cite{prillo2020softsort} and find that~\cite{prillo2020softsort} is overly dependent on temperature $\temperature$ to break out the ranks, and its gradients are too unreliable to train our model. 
            Hence, we stick with the hard sorting of $\score$ and $\overlapMat$.

        \subsubsection{NMS as a Matrix Operation}\label{subsec:matrix_formulation}
            The rescoring process of the~\classicalNms~is greedy set-based~\cite{prokudin2017learning} and only considers overlaps with unsuppressed boxes. 
            We first generalize this rescoring 
            by accounting for the effect of all (suppressed and unsuppressed) boxes as
            \begin{align}
                r_i &\approx \max\left(s_i - \sum\limits_{j=1}^{i-1}\pruneof{\overlap_{ij}}r_j,~0 \right)
                \label{eq:diff_nms_rescore}
            \end{align}
            using the relaxation of logical OR $\bigvee$ operator as $\sum$~\cite{van2020analyzing, li2019augmenting}.
            See \myReferSection{sec:NMS_explanation} of the supplementary material for an alternate explanation of \eqref{eq:diff_nms_rescore}.
            The presence of $r_j$ on the RHS of \eqref{eq:diff_nms_rescore} prevents suppressed boxes from influencing other boxes hugely. 
            When $\prune$ outputs discretely as $\{0, 1\}$ as in~\classicalNms, scores $s_i$ are guaranteed to be suppressed to $r_i=0$ or left unchanged $r_i=s_i$ thereby implying $r_i \leq s_i~\forall~i$. 
            We write the rescores $\rescore$ in a matrix formulation as
            \begin{align}
                \begin{bmatrix}
                r_1 \\
                r_2 \\
                r_3 \\
                \vdots \\
                r_n\\
                \end{bmatrix}
                    &\!\approx\! 
                \max\left(\begin{bmatrix}
                c_1\\
                c_2\\
                c_3\\
                \vdots \\
                c_n\\
                \end{bmatrix}
                ,
                \begin{bmatrix}
                0 \\
                0 \\
                0 \\
                \vdots \\
                0\\
                \end{bmatrix}
                \right),
            \end{align}
            with
            \begin{align}
                \begin{bmatrix}
                c_1\\
                c_2\\
                c_3\\
                \vdots \\
                c_n\\
                \end{bmatrix}
                &=
                \begin{bmatrix}
                s_1\\
                s_2\\
                s_3\\
                \vdots \\
                s_n\\
                \end{bmatrix}
                -
                \begin{bmatrix}
                0 & 0 & \dots & 0\\
                \pruneof{\overlap_{21}}\!&\!0\!&\!\dots\!&\!0\\
                \pruneof{\overlap_{31}}\!&\!\pruneof{\overlap_{32}}\!&\!\dots\!&\!0\\
                \vdots\!& \vdots & \vdots\!&\!\vdots \\
                \pruneof{\overlap_{n1}}\!&\!\pruneof{\overlap_{n2}} & \dots & 0 \\
                \end{bmatrix} 
                \begin{bmatrix}
                r_1 \\
                r_2 \\
                r_3 \\
                \vdots \\
                r_n\\
                \end{bmatrix}.
            \end{align}
            The above two equations are written compactly as 
            \begin{align}
                \rescore &\approx \max(\score - \pruneMat \rescore,\zeroVector),
                \label{eq:diff_nms_recursive}
            \end{align}
            where $\pruneMat$, called the Prune Matrix, is obtained 
            when the pruning function $\prune$ operates element-wise
            on $\overlapMatLower$. 
            Maximum operation makes \eqref{eq:diff_nms_recursive} non-linear~\cite{kumar2013estimation} and, thus, difficult to solve. 
            However, to avoid recursion, we use 
            \begin{align}
                \rescore &\approx \clip{\left( \identity + \pruneMat \right)^{-1}\score}, \label{eq:diff_nms_full}
            \end{align}
            as the solution to \eqref{eq:diff_nms_recursive} with $\identity$ being the identity matrix. 
            Intuitively, if the matrix inversion is considered division in \eqref{eq:diff_nms_full} and the boxes have overlaps, the rescores are the scores divided by a number greater than one and are, therefore, lesser than scores. 
            If the boxes do not overlap, the division is by one and rescores equal scores.
            
            Note that the $\identity + \pruneMat$ in~\eqref{eq:diff_nms_full} is a lower triangular matrix with ones on the principal diagonal. Hence, $\identity + \pruneMat$ is always full rank and, therefore, always invertible.

        \subsubsection{Grouping}
            We next observe that the object detectors output multiple boxes for an object, and a good detector outputs boxes wherever it finds objects in the monocular image. 
            Thus, we cluster the boxes in an image in an unsupervised manner based on \iouTwoD~overlaps to obtain the groups $\groups$. 
            Grouping thus mimics the grouping of the~\classicalNms, but does not rescore the boxes. 
            As clustering limits interactions to intra-group interactions among the boxes, we write~\eqref{eq:diff_nms_full} as
            \begin{align}
               \rescoreGroup &\approx \clip{\left( \identityGroup + \pruneMatGroup \right)^{-1}\scoreGroup}.
               \label{eq:diff_nms_group_intermediate}
            \end{align}
            This results in taking smaller matrix inverses in \eqref{eq:diff_nms_group_intermediate} than \eqref{eq:diff_nms_full}.
            
            We use a simplistic grouping algorithm, \thatIs, we form a group $\groupMember$ with boxes  having high \iouTwoD~overlap with the top-ranked box, given that we sorted the scores. 
            As the group size is limited by $\groupSize$, we choose a minimum of $\groupSize$ and the number of boxes in $\groupMember$. 
            We next delete all the boxes of this group and iterate until we run out of boxes.
            Also, grouping uses \iouTwoD~since we can achieve meaningful clustering in \twoD. 
            We detail this unsupervised grouping in Alg.~\ref{alg:group}. 

        \subsubsection{Masking}
            \classicalNmsCaps~considers the \iouTwoD~of the top-scored box with other boxes. 
            This consideration is equivalent to only keeping the column of $\overlapMat$ corresponding to the top box while assigning the rest of the columns to be zero. 
            We implement this through masking of $\pruneMatGroup$. 
            Let $\maskGroup$ denote the binary mask corresponding to group $\groupMember$. 
            Then, entries in the binary matrix $\maskGroup$ in the column corresponding to the top-scored box are $1$ and the rest are $0$.
            Hence, only one of the columns in $\maskGroup\elementMul~\pruneMatGroup$ is non-zero. 
            Now, $\identityGroup+\maskGroup\elementMul\pruneMatGroup$ is a Frobenius matrix (Gaussian transformation) and we, therefore, invert this matrix by simply subtracting the second term~\cite{golub2013matrix}. 
            In other words, $(\identityGroup+ \maskGroup\elementMul\pruneMatGroup)^{-1} =\identityGroup - \maskGroup\elementMul\pruneMatGroup$. Hence, we simplify \eqref{eq:diff_nms_group_intermediate} further to get
            \begin{align}
               \rescoreGroup &\approx \clip{ \left( \identityGroup - \maskGroup\elementMul\pruneMatGroup \right)\scoreGroup}.
               \label{eq:diff_nms_group_mask}
            \end{align}
            Thus, masking allows to bypass the computationally expensive matrix inverse operation altogether.
            
            We call the NMS based on \eqref{eq:diff_nms_group_mask} as Grouped Mathematically Differentiable Non-Maximal Suppression or \methodName. 
            We summarize the complete \methodName~in Alg.~\ref{alg:diff_nms} and show its block-diagram in \myReferFigure{fig:overview}(c). 
            \methodName~in \myReferFigure{fig:overview}(c) provides two gradients - one through $\score$ and other through $\overlapMat$.

        \subsubsection{Pruning Function}\label{sec:pruning}

            As explained in \myReferSection{subsec:notations}, the pruning function $\prune$ decides whether to keep the box in the final set of predictions $\boxesAfterNMS$ or not based on \iouTwoD~overlaps, \thatIs, $\pruneof{\overlap_i} = 1$ denotes the box $\boxi$ is suppressed while $\pruneof{\overlap_i} = 0$ denotes $\boxi$ is kept in $\boxesAfterNMS$. 
            
            \classicalNmsCaps~uses the threshold as the pruning function, which does not give useful gradients. 
            Therefore, we  considered three different functions for $\prune$: \basic, a temperature $(\temperature)$-controlled \exponentialPruning, and Sigmoidal function.
            \begin{compactitem}
                \item \textbf{Linear} Linear pruning function~\cite{bodla2017soft} is $\prune(\overlap)=\overlap$.
                \item \textbf{\exponentialPruning} \exponentialPruning~pruning function~\cite{bodla2017soft} is $\prune(\overlap) = 1- \mathrm{exp}\autoBraces{-\frac{\overlap^2}{\temperature}}$.
                \item \textbf{Sigmoidal} Sigmoidal pruning function is $\prune(\overlap) = \sigma\autoBraces{\frac{\overlap-\nmsThresh}{\temperature}}$ with $\sigma$ denoting the standard sigmoid. Sigmoidal function appears as the binary cross entropy relaxation of the subset selection problem~\cite{paulus2020gradient}.
            \end{compactitem}
            
\begin{figure}[!tb]
                \centering
                \includegraphics[width=0.99\linewidth]{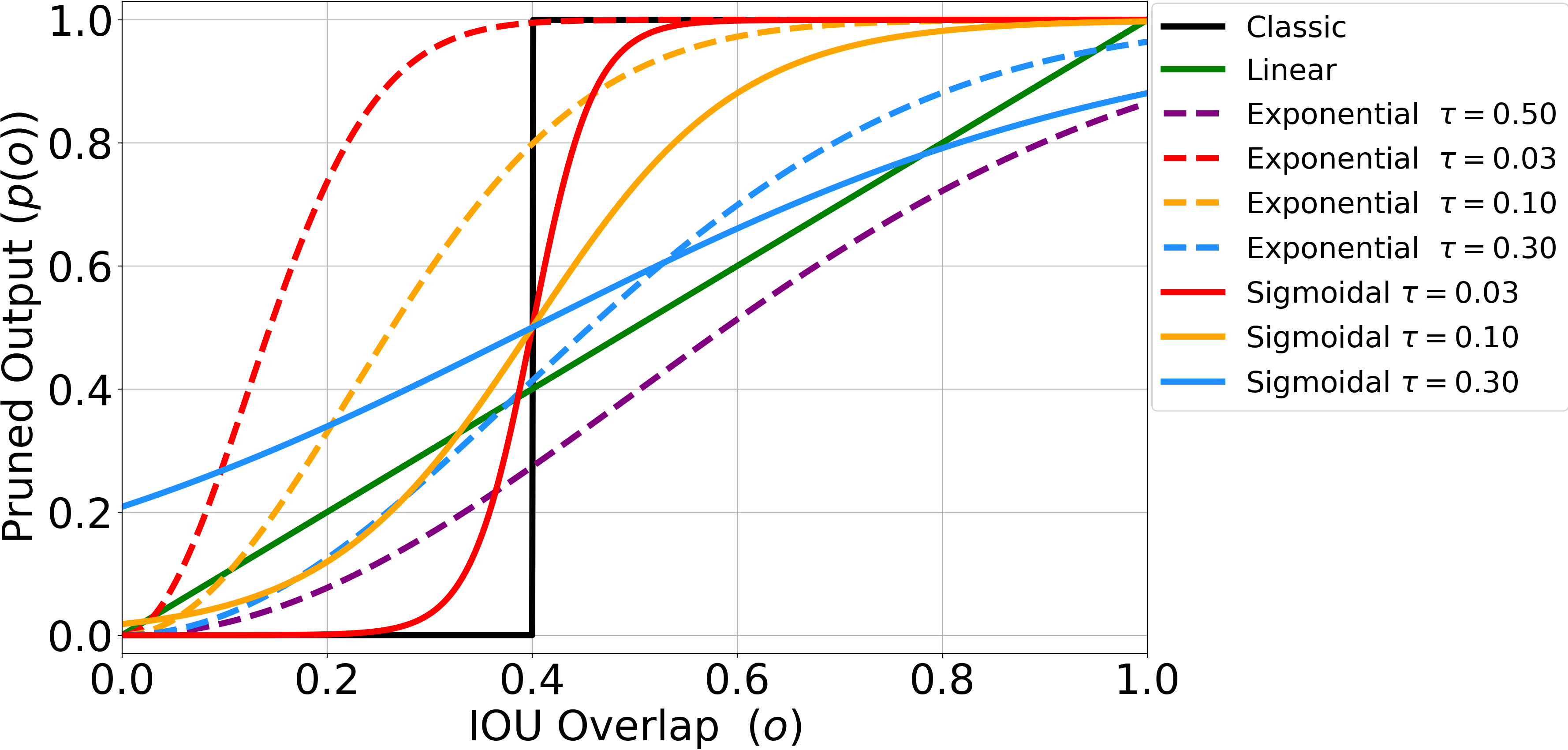}
                \vspace{1mm}
                \caption{\textbf{Pruning functions} $\prune$ of the \classicalNmsShort~and \methodName. We use the Linear and Exponential pruning of the \softNmsCaps~\cite{bodla2017soft} while training with the \methodName.}
                \label{fig:pruning}
            \end{figure}

            We show these pruning functions in \myReferFigure{fig:pruning}. 
            The ablation studies (\myReferSection{sec:results_ablation}) show that choosing $\prune$ as \basic~yields the simplest and the best \methodName.

    \subsection{Differences from Existing NMS}\label{sec:difference}

        Although no differentiable NMS has been proposed for the monocular \threeD~object detection, we compare our \methodName~with the NMS proposed for \twoD~object detection,~\twoD~pedestrian detection, \twoD~salient object detection, and~\threeD~object detection in \myReferTable{tab:nms_overview}.
        No method described in \myReferTable{tab:nms_overview} has a matrix-based closed-form mathematical expression of the NMS.
        ~\classicalNmsShortCaps, \softNmsShortCaps~\cite{bodla2017soft} and \distanceNmsCaps~\cite{shi2020distance} are used at the inference time, while \methodName~is used during both training and inference.
        \distanceNmsCaps~\cite{shi2020distance} updates the $z$-coordinate of the box after NMS as the weighted average of the $z$-coordinates of top-$\kappa$ boxes.
        QUBO-NMS~\cite{rujikietgumjorn2013optimized}, Point-NMS~\cite{lee2016individualness, some2020determinantal}, and MAP-NMS~\cite{zhang2016unconstrained} are not used in end-to-end training.
        \cite{azadi2017learning} proposes a trainable Point-NMS.
        The Structured-SVM based NMS~\cite{desai2011discriminative, wan2015end} rely on structured SVM to obtain the rescores. 
        Adaptive-NMS\cite{liu2019adaptive} uses a separate neural network to predict the \classicalNms~threshold~$\nmsThresh$.
        The trainable neural network based NMS (NN-NMS)~\cite{hosang2016convnet, prokudin2017learning, hosang2017learning} use a separate neural network containing multiple layers and/or message-passing to approximate the NMS and do not use the pruning function.
        Unlike these methods, \methodName~uses a single layer and does not require multiple layers or message passing. Our NMS is parallel up to group (denoted by $\groups$). 
        However, $\numberElem{\groups}$ is, in general, $<< \numberElem{\boxes}$ in the NMS. 

        \begin{table}[!tb]
            \caption{Overview of different NMS. [Key: Train= End-to-end Trainable, Prune= Pruning function, \#Layers= Number of layers, Par= Parallelizable]}
            \label{tab:nms_overview}
            \setlength{\tabcolsep}{0.03cm}
            \centering
            \footnotesize
            \begin{tabular}{lcccccc}
                \myTopRule
                \textbf{NMS}                                            & \textbf{Train} & \textbf{Rescore} & \textbf{Prune}  & \textbf{\#Layers} & \textbf{Par} & \\
                \myTopRule
                \classicalNmsShortCaps~                                       & \xmark         & \xmark           & Hard            & -          & $\bigO{\numberElem{\groups}}$ \\
                \softNmsCaps~\cite{bodla2017soft}                             & \xmark         & \xmark           & Soft            & -          & $\bigO{\numberElem{\groups}}$ \\
                \distanceNmsCaps~\cite{shi2020distance}                       & \xmark         & \xmark           & Hard            & -          & $\bigO{\numberElem{\groups}}$ \\
                QUBO-NMS~\cite{rujikietgumjorn2013optimized}                  & \xmark         & Optimization     & \xmark          & -          & -\\
                Point-NMS~\cite{lee2016individualness, some2020determinantal} & \xmark         & Point Process    & \xmark          & -          & -\\
                Trainable Point-NMS~\cite{azadi2017learning}                  & \cmark         & Point Process    & \xmark          & -          & -\\
                MAP-NMS~\cite{zhang2016unconstrained}                         & \xmark         & MAP              & \xmark          & -          & -\\
                Structured-NMS~\cite{desai2011discriminative, wan2015end}     & \xmark         & SSVM             & \xmark          & -          & -\\
                Adaptive-NMS~\cite{liu2019adaptive}                           & \xmark         & \xmark           & Hard            & $\!>\!1$  & $\bigO{\numberElem{\groups}}$\\
                NN-NMS~\cite{hosang2016convnet, prokudin2017learning, hosang2017learning} & \cmark & Neural Network    & \xmark      & $\!>\!1$ & $\bigO{1}$ \\
                \bottomrule
                \!\methodName~(Ours)                                          & \cmark         & Matrix           & Soft           & $1$         & $\bigO{\numberElem{\groups}}$ \\
                \myTopRule
            \end{tabular}
        \end{table}

    \subsection{Target Assignment and Loss Function}\label{sec:target_loss}
    
        \noindent\textbf{Target Assignment.}
            Our method consists of M3D-RPN~\cite{brazil2019m3d} and uses binning and self-balancing confidence~\cite{brazil2020kinematic}. The boxes' self-balancing confidence are used as scores $\score$, which pass through the \methodName~layer to obtain the rescores $\rescore$. The rescores signal the network if the \emph{best} box has not been selected for a particular object. 
            
            We extend the notion of the best \twoD~box~\cite{prokudin2017learning} to \threeD. The best box has the highest product of \iouTwoD~and \giouThreeD~\cite{rezatofighi2019generalized} with ground truth $\groundTruthj$. If the product is greater than a certain threshold $\beta$, it is assigned a positive label. Mathematically,
            \begin{align}
                \text{target}(\boxi) &= \left\{\begin{array}{@{}ll@{}}
                    \multirow{2}{*}{$1$,} & \text{if}~\exists~\groundTruthj~\text{st}~i=\argmax q(\boxj, \groundTruthj)\\ 
                                          & \phantom{\text{if}~\exists~\groundTruthj~s.~t.~i= }\text{and}~q(\boxi, \groundTruthj) \ge \beta\\
                    0, & \text{otherwise}
                    \end{array}\right.
            \end{align}
            with $q(\boxj, \groundTruthj) = \text{\iouTwoD}(\boxj,\groundTruthj)~\left(\frac{1+\text{\giouThreeD}(\boxj,\groundTruthj)}{2}\right)$. 
            \giouThreeD~is known to provide signal even for non-intersecting boxes~\cite{rezatofighi2019generalized}, where the usual \iouThreeD~is always zero. 
            Therefore, we use \giouThreeD~instead of regular \iouThreeD~for figuring out the best box in \threeD~as many \threeD~boxes have a zero \iouThreeD~overlap with the ground truth.
            For calculating \giouThreeD, we first calculate the volume $\vol$ and hull volume $\vol_{hull}$ of the \threeD~boxes.  $\vol_{hull}$ is the product of \giouTwoD~in Birds Eye View (BEV), removing the rotations and hull of the $Y$ dimension. \giouThreeD~is then given by
            \begin{align}
                \text{\giouThreeD}(\boxi,\boxj) &= \frac{\vol(\boxi \cap \boxj)}{\vol(\boxi \cup \boxj)} + \frac{\vol(\boxi \cup \boxj)}{\vol_{hull}(\boxi, \boxj)} - 1.
            \end{align} 
            
        \noindent\textbf{Loss Function.}
            Generally the number of best boxes is less than the number of ground truths in an image, as there could be some ground truth boxes for which no box is predicted. The tiny number of best boxes introduces a far-heavier skew than the foreground-background classification. Thus, we use the modified \aploss~\cite{chen2020ap} as our loss after NMS since \aploss~does not suffer from class imbalance~\cite{chen2020ap}. 
            
            Vanilla \aploss~treats boxes of all images in a mini-batch equally, and the gradients are back-propagated through all the boxes. We remove this condition and rank boxes in an image-wise manner. In other words, if the best boxes are correctly ranked in one image and are not in the second, then the gradients only affect the boxes of the second image. We call this modification of \aploss~
            as \emph{\imageWise} \aploss. In other words,
            \vspace{-0.15cm}
            \begin{align}
                \loss_{\imageWise} = \frac{1}{N}\sum_{m=1}^N \apMath(\rescore^{(m)}, \text{target}(\boxes^{(m)})),
            \end{align}
            where $\rescore^{(m)}$ and $\boxes^{(m)}$ denote the rescores and the boxes of the $m^{\text{th}}$ image in a mini-batch respectively. This is different from previous NMS approaches~\cite{hosang2016convnet, henderson2016end, prokudin2017learning, hosang2017learning}, which use classification losses. Our ablation studies (\myReferSection{sec:results_ablation}) show that the \imageWise~\aploss~is better suited to be used after NMS than the classification loss.
    
            Our overall loss function is thus given by $\loss = \lossBefore + \lossWeigh \lossAfter$ where $\lossBefore$ denotes the losses before the NMS including classification, \twoD~and \threeD~regression as well as confidence losses, and $\lossAfter$ denotes the loss term after the NMS, which is the \imageWise~\aploss~with $\lossWeigh$ being the weight. 
            See \myReferSection{sec:appendix_loss} of the supplementary material for more details of the loss function.

\section{Experiments}\label{sec:experiments}

    Our experiments use the most widely used \kitti~autonomous driving dataset~\cite{geiger2012we}. 
    We modify the publicly-available PyTorch~\cite{paszke2019pytorch} code of Kinematic-3D~\cite{brazil2020kinematic}.
    \cite{brazil2020kinematic} uses DenseNet-121~\cite{huang2017densely} trained on ImageNet as the backbone and $n_h\!=\!1{,}024$ using \threeD-RPN settings of~\cite{brazil2019m3d}. 
    As \cite{brazil2020kinematic} is a video-based method while \methodName~is an image-based method, we use the best image model of~\cite{brazil2020kinematic} henceforth called~\kinematicImage~as our baseline for a fair comparison.
    \kinematicImage~is built on M3D-RPN~\cite{brazil2019m3d} and uses binning and self-balancing confidence. 

    \noindent\textbf{Data Splits.}
        There are three commonly used data splits of the \kitti~dataset; we evaluate our method on all three.
    
        \textit{Test Split}: Official \kitti~\threeD~benchmark~\cite{kitti2012benchmark}
        consists of $7{,}481$ training and $7{,}518$ testing images~\cite{geiger2012we}.
        
        \textit{\valOne~Split}: It partitions the $7{,}481$ training images into $3{,}712$ training and $3{,}769$ validation images~\cite{chen20153d,simonelli2020disentangling, brazil2020kinematic}.
        
        \textit{\valTwo~Split}: It partitions the $7{,}481$ training images into $3{,}682$ training and $3{,}799$ validation images~\cite{xiang2017subcategory}.
    
    \noindent\textbf{Training.}
        Training is done in two phases - warmup and full~\cite{brazil2020kinematic}. 
        We initialize the model with the confidence prediction branch from warmup weights and finetune using the self-balancing loss~\cite{brazil2020kinematic} and \imageWise~\aploss~\cite{chen2020ap} after our \methodName. 
        See \myReferSection{sec:training_additional} of the supplementary material for more training details. 
        We keep the weight $\lossWeigh$ at $0.05$. 
        Unless otherwise stated, we use $\prune$ as the \basic~function (this does not require $\temperature$) with $\groupSize=100$. 
        $\nmsThresh, \validBoxThresh$ and $\beta$ are set to $0.4$~\cite{brazil2019m3d, brazil2020kinematic}, $0.3$ and $0.3$ respectively.

    \noindent\textbf{Inference.}
        We multiply the class and predicted confidence to get the box's overall score in inference as in~\cite{tychsen2018improving, shi2020distance, kim2020probabilistic}. 
        See \myReferSection{sec:results_kitti_val1} for training and inference times.

        \begin{table}[!tb]
            \caption{\apThreeDForty~and \apBevForty~comparisons on the \kitti~Test Cars (\iouThreeD~$\geq 0.7$). Previous results are quoted from the official leader-board or from papers.[Key: \firstkey{Best}, \secondkey{Second Best}].
            }
            \label{tab:results_kitti_test}
            \centering
            \footnotesize
            \setlength\tabcolsep{2.00pt}
            \begin{tabular}{tl m ccc  m ccct}
                \myTopRule
                \addlinespace[0.01cm]
                \multirow{2}{*}{Method} & \multicolumn{3}{cm}{\apThreeDForty ($\uparrow$)} & \multicolumn{3}{ct}{\apBevForty ($\uparrow$)}\\ 
                & Easy & Mod & Hard & Easy & Mod & Hard\\ 
                \myTopRule
                FQNet~\cite{liu2019deep}                   &   $2.77$       &   $1.51$       &   $1.01$      &   $5.40$       &   $3.23$       &   $2.46$      \\
                ROI-10D~\cite{manhardt2019roi}             &   $4.32$       &   $2.02$       &   $1.46$      &   $9.78$       &   $4.91$       &   $3.74$      \\
                GS3D~\cite{li2019gs3d}                     &   $4.47$       &   $2.90$       &   $2.47$      &   $8.41$       &   $6.08$       &   $4.94$      \\
                MonoGRNet~\cite{qin2019monogrnet}          &   $9.61$       &   $5.74$       &   $4.25$      & $18.19$        &   $11.17$      &   $8.73$      \\
                MonoPSR~\cite{ku2019monocular}             &  $10.76$       &   $7.25$       &   $5.85$      & $18.33$        & $12.58$        &   $9.91$      \\
                MonoDIS~\cite{simonelli2019disentangling}  &  $10.37$       &   $7.94$       &   $6.40$      & $17.23$        & $13.19$        & $11.12$       \\
                UR3D~\cite{shi2020distance}                &  $15.58$       &   $8.61$       &   $6.00$      & $21.85$        & $12.51$        & $9.20$        \\
                M3D-RPN~\cite{brazil2019m3d}               &  $14.76$       &   $9.71$       &   $7.42$      & $21.02$        & $13.67$        & $10.23$       \\
                SMOKE~\cite{liu2020smoke}                  & $14.03$        &   $9.76$       & $7.84$        & $20.83$        & $14.49$        & $12.75$       \\
                MonoPair~\cite{chen2020monopair}           & $13.04$        &   $9.99$       & $8.65$        & $19.28$        & $14.83$        & $12.89$       \\
                RTM3D~\cite{li2020rtm3d}                   & $14.41$        &   $10.34$      & $8.77$        & $19.17$        & $14.20$        & $11.99$       \\
                AM3D~\cite{ma2019accurate}                 & $16.50$        &   $10.74$      & \second{9.52}  & $25.03$        & 17.32        & \first{14.91} \\
                MoVi-3D~\cite{simonelli2020towards}        & $15.19$        &   $10.90$      & $9.26$        & $22.76$        & $17.03$        & $10.86$       \\
                RAR-Net~\cite{liu2020reinforced}           & $16.37$        &   $11.01$      & \second{9.52} & $22.45$        & $15.02$        & $12.93$       \\
                M3D-SSD~\cite{luo2021m3dssd}               & $17.51$        &   $11.46$      & 8.98          & $24.15$        & $15.93$        & $12.11$       \\
                DA-3Ddet~\cite{ye2020monocular}            & $16.77$        &   $11.50$      & 8.93          & -              & -              & -             \\
                D4LCN~\cite{ding2020learning}              & $16.65$        & $11.72$        & 9.51          & $22.51$        & $16.02$        &  $12.55$      \\
                \kinematicVideo~\cite{brazil2020kinematic} & \first{19.07}  & \first{12.72}  & $9.17$        & \first{26.69}  & \second{17.52}  & 13.10\\
                \hline
                \methodName~(Ours) & \second{18.10} & \second{12.32} & \first{9.65} & \second{26.19} & \first{18.27} & \second{14.05}\\
                \myTopRule
            \end{tabular}
        \end{table}    
    
    \noindent\textbf{Evaluation Metrics.} 
        \kitti~uses \apThreeDForty~metric to evaluate object detection following~\cite{simonelli2019disentangling, simonelli2020disentangling}.
        \kitti~benchmark evaluates on three object categories: Easy, Moderate and Hard. 
        It assigns each object to a category based on its occlusion, truncation, and height in the image space. 
        The \apThreeDForty~performance on the Moderate category compares different models in the benchmark~\cite{geiger2012we}. 
        We focus primarily on the Car class following~\cite{brazil2020kinematic}.

    \subsection{\kitti~Test 3D~Object Detection}\label{sec:results_kitti_test}
        \myReferTable{tab:results_kitti_test} summarizes the results of \threeD~object detection and BEV evaluation on \kitti~Test Split. The results in \myReferTable{tab:results_kitti_test} show that \methodName~outperforms the baseline M3D-RPN~\cite{brazil2019m3d} by a significant margin and several other \sota~methods on both the tasks.~\methodName~also outperforms augmentation based approach MoVi-3D~\cite{simonelli2020towards} and depth-convolution based D4LCN~\cite{ding2020learning}. 
        Despite being an image-based method, \methodName~performs competitively to the video-based method \kinematicVideo~\cite{brazil2020kinematic}, outperforming it on the most-challenging Hard set.
        \begin{table*}[t]
            \caption{\apThreeDForty~and \apBevForty~comparisons on \kitti~\valOne~Cars. [Key: \firstkey{Best}, \secondkey{Second Best}].
            }
            \label{tab:results_kitti_val1}
            \centering
            \footnotesize
            \setlength\tabcolsep{0.1cm}
            \begin{tabular}{tl m ccc t ccc m ccc t ccct}
                \myTopRule
                \addlinespace[0.01cm]
                \multirow{3}{*}{Method} & \multicolumn{6}{cm}{\iouThreeD~$\geq 0.7$} & \multicolumn{6}{ct}{\iouThreeD~$\geq 0.5$}\\\cline{2-13}
                & \multicolumn{3}{ct}{\apThreeDForty ($\uparrow$)} & \multicolumn{3}{cm}{\apBevForty ($\uparrow$)} & \multicolumn{3}{ct}{\apThreeDForty ($\uparrow$)} & \multicolumn{3}{ct}{\apBevForty ($\uparrow$)}\\
                & Easy & Mod & Hard & Easy & Mod & Hard & Easy & Mod & Hard & Easy & Mod & Hard\\ 
                \myTopRule
                MonoDR~\cite{beker2020monocular}                                 & $12.50$        & $7.34$         & $4.98$        & $19.49$        & $11.51$        & $8.72$       & -              & -              & -              & -              & -              & -             \\
                MonoGRNet~\cite{qin2019monogrnet} in~\cite{chen2020monopair}             & $11.90$        & $7.56$         & $5.76$        & $19.72$        & $12.81$        & $10.15$       & $47.59$        & $32.28$        & $25.50$        & $52.13$        & $35.99$        & $28.72$\\
                MonoDIS~\cite{simonelli2019disentangling} in~\cite{simonelli2020disentangling} & $11.06$        & $7.60$         & $6.37$        & $18.45$        & $12.58$        & $10.66$       & -              & -              & -              & -              & -              & -             \\
                M3D-RPN~\cite{brazil2019m3d} in~\cite{brazil2020kinematic}               & $14.53$        & $11.07$        & $8.65$        & $20.85$        & $15.62$        & $11.88$       & $48.56$        & $35.94$        & $28.59$        & $53.35$        & $39.60$        & $31.77$       \\
                MoVi-3D~\cite{simonelli2020towards}                                          & $14.28$        & $11.13$        & $9.68$        & $22.36$        & $17.87$        & $15.73$       & -              & -              & -              & -              & -              & -             \\
                MonoPair~\cite{chen2020monopair}                                         & $16.28$        & $12.30$        & $10.42$       & $24.12$        & $18.17$        & \second{15.76}& $55.38$        & \first{42.39}  & \first{37.99}  & $61.06$        & \first{47.63}        & \first{41.92}       \\
                \kinematicImage~\cite{brazil2020kinematic}                              & $18.28$        & $13.55$        & $10.13$       & $25.72$        & $18.82$        & $14.48$       & $54.70$        & $39.33$        & $31.25$        & $60.87$        & $44.36$        & $34.48$       \\
                \kinematicVideo~\cite{brazil2020kinematic}                             & \first{19.76}  & \second{14.10} & \second{10.47} &  \first{27.83} & \second{19.72} & $15.10$       & \second{55.44} & $39.47$        & $31.26$        & \second{61.79} & $44.68$        & $34.56$\\
                \hline
                \methodName~(Ours)                                                       & \second{19.67} & \first{14.32}  & \first{11.27} & \second{27.38} & \first{19.75}  & \first{15.92} & \first {55.62} & \second{41.07} & \second {32.89}& \first {61.83} & \second{44.98} & \second{36.29}\\
                \myTopRule
            \end{tabular}
            \vspace{-0.2cm}
        \end{table*}            
    
        \begin{table}[t]
            \caption{\apThreeDForty~and \apBevForty~comparisons with other NMS on \kitti~\valOne~Cars (\iouThreeD~$\geq 0.7$). [Key: C= \classicalNmsShortCaps, S= \softNmsCaps\cite{bodla2017soft}, D= \distanceNmsCaps\cite{shi2020distance}, G= \methodName]}
            \label{tab:results_kitti_val1_other_nms}
            \centering
            \footnotesize
            \setlength\tabcolsep{0.075cm}
            \begin{tabular}{tl t c m ccc t ccct}
                \myTopRule
                \addlinespace[0.01cm]
                \multirow{2}{*}{Method}& \multirow{2}{*}{\shortstack{Infer\\NMS}} & \multicolumn{3}{ct}{\apThreeDForty ($\uparrow$)} & \multicolumn{3}{ct}{\apBevForty ($\uparrow$)}\\
                & & Easy & Mod & Hard & Easy & Mod & Hard\\ 
                \myTopRule
                \kinematicImage      &C& $18.28$        & $13.55$        & $10.13$       & $25.72$        & $18.82$        & $14.48$\\
                \kinematicImage      &S& $18.29$        & $13.55$        & $10.13$       & $25.71$        & $18.81$        & $14.48$\\
                \kinematicImage      &D& $18.25$        & $13.53$        & $10.11$       & $25.71$        & $18.82$        & $14.48$\\
                \kinematicImage      &G& $18.26$        & $13.51$        & $10.10$       & $25.67$        & $18.77$        & $14.44$\\
                \hline
                \methodName~         &C& $19.67$ & $14.31$  & $11.27$ & $27.38$ & $19.75$  & $15.93$\\
                \methodName~         &S& $19.67$ & $14.31$  & $11.27$ & $27.38$ & $19.75$  & $15.93$\\
                \methodName~         &D& $19.67$ & $14.31$  & $11.27$ & $27.38$ & $19.75$  & $15.93$\\
                \methodName~         &G& $19.67$ & $14.32$  & $11.27$ & $27.38$ & $19.75$  & $15.92$\\
                \myTopRule
            \end{tabular}
            \vspace{0.2cm}
        \end{table}

    \subsection{\kitti~\valOne~3D~Object Detection}\label{sec:results_kitti_val1}
        
        \noindent\textbf{Results.}
            \myReferTable{tab:results_kitti_val1} summarizes the results of \threeD~object detection and BEV evaluation on \kitti~\valOne~Split at two \iouThreeD~thresholds of $0.7$ and $0.5$~\cite{chen2020monopair,brazil2020kinematic}. 
            The results in \myReferTable{tab:results_kitti_val1} show that \methodName~outperforms the baseline of M3D-RPN~\cite{brazil2019m3d} and \kinematicImage~\cite{brazil2020kinematic} by a significant margin. 
            Interestingly, \methodName~(an image-based method) also outperforms the video-based method \kinematicVideo~\cite{brazil2020kinematic} on most of the metrics. 
            Thus, \methodName~performs best on $6$ out of the $12$ cases ($3$ categories $\times~2$~tasks $\times~2$~thresholds) while second-best on all other cases. 
            The performance is especially impressive since the biggest improvements are shown on the Moderate and Hard set, where objects are more distant and occluded.

        \begin{figure}[!tb]
            \centering
            \begin{subfigure}{.5\linewidth}
              \centering
              \includegraphics[width=\linewidth]{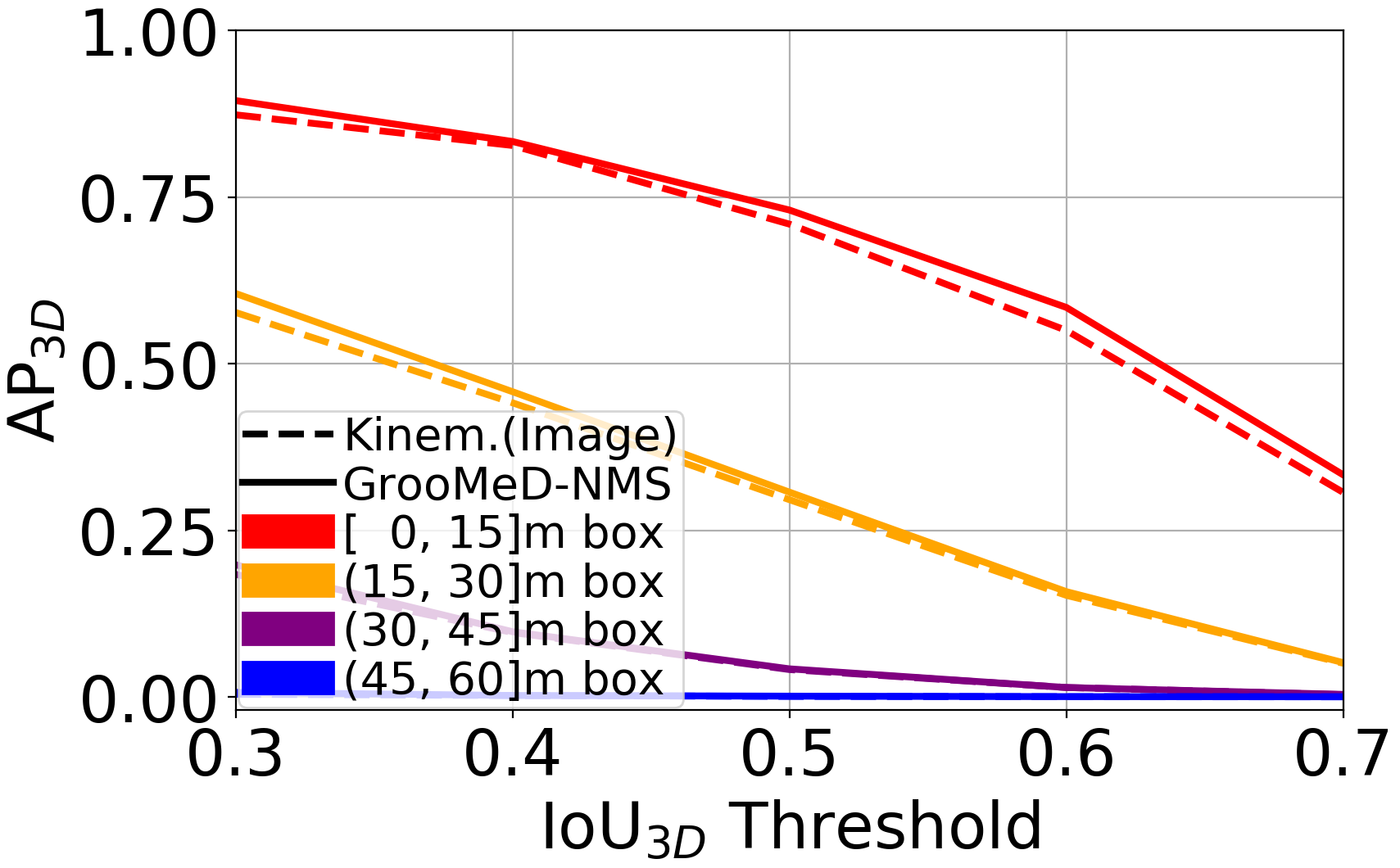}
              \caption{Linear Scale}
            \end{subfigure}%
            \begin{subfigure}{.5\linewidth}
              \centering
              \includegraphics[width=\linewidth]{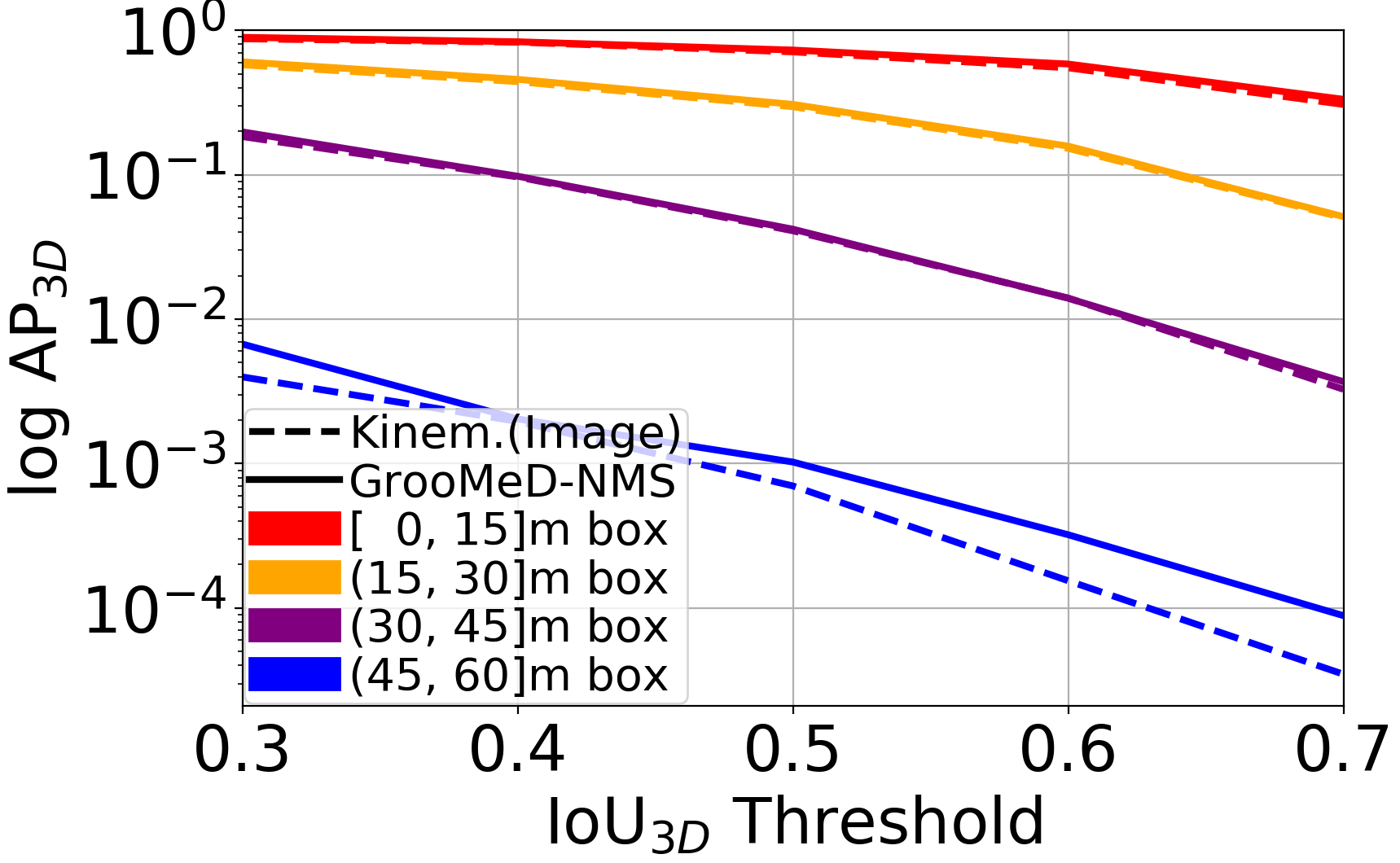}
              \caption{Log Scale}
            \end{subfigure}
            \caption{Comparison of \apthreeD~at different depths and \iouThreeD~matching thresholds on \kitti~\valOne~Split.}
            \label{fig:ap_ground_truth_threshold}
        \end{figure}

        \noindent\textbf{\apthreeD~at different depths and \iouThreeD~thresholds.}
            We next compare the \apthreeD~performance of \methodName~and \kinematicImage~on linear and log scale for objects at different depths of $[15,30, 45, 60]$ meters and \iouThreeD~matching criteria of $0.3\!\rightarrowRHD\!0.7$ in \myReferFigure{fig:ap_ground_truth_threshold} as in~\cite{brazil2020kinematic}. 
            \myReferFigure{fig:ap_ground_truth_threshold} shows that~\methodName~outperforms the \kinematicImage~\cite{brazil2020kinematic} at all depths and all \iouThreeD~thresholds.
    
        \noindent\textbf{Comparisons with other NMS.}
            We compare with the 
            \classicalNms, \softNmsCaps~\cite{bodla2017soft} and \distanceNmsCaps~\cite{shi2020distance} in \myReferTable{tab:results_kitti_val1_other_nms}.
            More detailed results are in \myReferTable{tab:results_kitti_val1_other_nms_detailed} of the supplementary material.
            The results show that NMS inclusion in the training pipeline benefits the performance, unlike~\cite{bodla2017soft}, which suggests otherwise. 
            Training with \methodName~helps because the network gets an additional signal through the \methodName~layer whenever the best-localized box corresponding to an object is not selected. 
            Interestingly, \myReferTable{tab:results_kitti_val1_other_nms} also suggests that replacing \methodName~with the~\classicalNms~in inference does not affect the performance.

        \begin{table*}[!tb]
            \caption{\apThreeDForty~and \apBevForty~comparisons on \kitti~\valTwo~Cars. [Key: \firstkey{Best}, *= Released, $^\dagger$= Retrained].
            }
            \label{tab:results_kitti_val2}
            \centering
            \footnotesize
            \setlength\tabcolsep{0.1cm}
            \begin{tabular}{tl m ccc t ccc m ccc t ccct}
                \myTopRule
                \addlinespace[0.01cm]
                \multirow{3}{*}{Method} & \multicolumn{6}{cm}{\iouThreeD~$\geq 0.7$} & \multicolumn{6}{ct}{\iouThreeD~$\geq 0.5$}\\\cline{2-13}
                & \multicolumn{3}{ct}{\apThreeDForty ($\uparrow$)} & \multicolumn{3}{cm}{\apBevForty ($\uparrow$)} & \multicolumn{3}{ct}{\apThreeDForty ($\uparrow$)} & \multicolumn{3}{ct}{\apBevForty ($\uparrow$)}\\
                & Easy & Mod & Hard & Easy & Mod & Hard & Easy & Mod & Hard & Easy & Mod & Hard\\ 
                \myTopRule
                M3D-RPN~\cite{brazil2019m3d}*               &$14.57$ &$10.07$ &$7.51$ &$21.36$ &$15.22$ &$11.28$ &$49.14$ &$34.43$ &$26.39$ &$53.44$ &$37.79$ &$29.36$\\
                \kinematicImage~\cite{brazil2020kinematic}$^\dagger$      &$13.54$        & $10.21$        & $7.24$        & $20.60$        & $15.14$        & $11.30$        & $51.53$        & $36.55$        & $28.26$        & $56.20$        & $40.02$        & $31.25$\\
                \hline
                \methodName~(Ours)                     & \first{14.72} & \first{10.87} & \first{7.67} & \first{22.03} & \first{16.05} & \first{11.93} & \first{51.91} & \first{36.78} & \first{28.40} & \first{56.29} & \first{40.31} & \first{31.39}
                \\
                \myTopRule
            \end{tabular}
        \end{table*}
        \begin{table*}[!tb]
            \caption{Ablation studies of our method on~\kitti~\valOne~Cars.}
            \label{tab:ablation}
            \centering
            \footnotesize
            \setlength{\tabcolsep}{0.08cm}
            \begin{tabular}{tc|lm ccc t ccc m ccc t ccct}
                \myTopRule
                \addlinespace[0.01cm]
                \multicolumn{2}{tcm}{\textbf{Change from \methodName~model:}} & \multicolumn{6}{cm}{\iouThreeD~$\geq 0.7$} & \multicolumn{6}{ct}{\iouThreeD~$\geq 0.5$}\\\cline{1-14}
                \multirow{2}{*}{Changed} & \multirow{2}{*}{From $\longrightarrowRHD$To} & \multicolumn{3}{ct}{\apThreeDForty ($\uparrow$)} & \multicolumn{3}{cm}{\apBevForty ($\uparrow$)} & \multicolumn{3}{ct}{\apThreeDForty ($\uparrow$)} & \multicolumn{3}{ct}{\apBevForty ($\uparrow$)}\\
                && Easy & Mod & Hard & Easy & Mod & Hard & Easy & Mod & Hard & Easy & Mod & Hard\\
                \myTopRule
                \multirow{3}{*}{Training}
                & Conf+NMS$\rightarrowRHD$No Conf+No NMS              &$16.66$ &$12.10$ &$9.40$  & $23.15$ &$17.43$ &$13.48$ &$51.47$ &$38.58$ &$30.98$ &$56.48$ &$42.53$ &$34.37$\\
                & Conf+NMS$\rightarrowRHD$Conf+No NMS                 &$19.16$ &$13.89$ &$10.96$ &	$27.01$ &$19.33$ &$14.84$ &$57.12$ &$41.07	$ &$32.79$ &$61.60$ &$44.58$ &$35.97$\\
                & Conf+NMS$\rightarrowRHD$No Conf+NMS                 &$15.02$ &$11.21$ &$8.83$ &$21.07$ &$16.27$ &$12.77$ &$48.01$ &$36.18$ &$29.96$ &$53.82$ &$40.94$ &$33.35$\\%
                \hline
                Initialization
                & No Warmup                                         &$15.33$ &$11.68$ &$8.78$ &$21.32$ &$16.59$ &$12.93$ &$49.15$ &$37.42$ &$30.11$ &$54.32$ &$41.44$ &$33.48$\\%
                \hline
                \multirow{4}{*}{\shortstack{Pruning\\Function}}
                & \basic$\rightarrowRHD$\exponentialPruning, $\temperature=1$     &$12.81$ &$9.26$ &$7.10$ &$17.07$ &$12.17$ &$9.25$ &$29.58$ &$20.42$ &$15.88$ &$32.06$ &$22.16$ &$17.20$\\
                & \basic$\rightarrowRHD$\exponentialPruning, $\temperature=0.5$\cite{bodla2017soft}   &$18.63$ &$13.85$ &$10.98$ &$27.52	$ &$20.14$ &$15.76$ &$56.64$ &$41.01$ &$32.79$ &$61.43$ &$44.73$ &$36.02$\\
                & \basic$\rightarrowRHD$\exponentialPruning, $\temperature=0.1$   &$18.34$ &$13.79$ &$10.88$ &$27.26$ &$19.71$ &$15.90$ &$56.98$ &$41.16$ &$32.96$ &$62.77$ &$45.23$ &$36.56$\\
                &\basic$\rightarrowRHD$Sigmoidal, $\temperature=0.1$ &$17.40$ &$13.21$ &$9.80$ &$26.77$ &$19.26$ &$14.76$ &$55.15$ &$40.77$ &$32.63$ &$60.56$ &$44.23$ &$35.74$\\
                \hline 
                \multirow{2}{*}{Group+Mask}
                & Group+Mask$\rightarrowRHD$No Group              &$18.43$ &$13.91$ &$11.08$ &$26.53$ &$19.46$ &$15.83$ &$55.93$ &$40.98$ &$32.78$ &$61.02$ &$44.77$ &$36.09$\\%
                & Group+Mask$\rightarrowRHD$Group+No Mask        &$18.99$ &$13.74$ &$10.24$ &$26.71$ &$19.21$ &$14.77$ &$55.21$ &$40.69$ &$32.55$ &$61.74$ &$44.67$ &$36.00$    \\%
                \hline
                \multirow{2}{*}{Loss}
                & \imageWise~\ap$\rightarrowRHD$Vanilla~\ap          &$18.23$ &$13.73$ &$10.28$ &$26.42$ &$19.31$ &$14.76$ &$54.47$ &$40.35$ &$32.20$ &$60.90$ &$44.08$ &$35.47$    \\%
                & \imageWise~\ap$\rightarrowRHD$BCE                    &$16.34$ &$12.74$ &$ 9.73$ &$22.40$ &$17.46$ &$13.70$ &$52.46$ &$39.40$ &$31.68$ &$58.22$ &$43.60$ &$35.27$            \\%
                \hline
                Inference
                & Class*Pred$\rightarrowRHD$Class                   &$18.26$& $13.36$& $10.49$& $25.39$& $18.64$& $15.12$& $52.44$& $38.99$& $31.3$& $57.37$& $42.89$& $34.68$\\%
                NMS Scores
                & Class*Pred$\rightarrowRHD$Pred                    & $17.51$& $12.84$& $9.55$& $24.55$& $17.85$& $13.63$& $52.78$& $37.48$& $29.37$& $58.30$& $41.26$& $32.66$\\%
                \hline
                {---} & \textbf{\methodName~(best model)}     & $\mathbf{19.67}$ & $\mathbf{14.32}$  & $\mathbf{11.27}$ & $\mathbf{27.38}$ & $\mathbf{19.75}$  & $\mathbf{15.92}$ & $\mathbf {55.62}$ & $\mathbf{41.07}$ & $\mathbf {32.89}$& $\mathbf {61.83}$ & $\mathbf{44.98}$ & $\mathbf{36.29}$\\
                \myTopRule
            \end{tabular}
            \vspace{-0.4cm}
        \end{table*}
        
\begin{figure}[t]
            \centering
            \includegraphics[width=0.67\linewidth]{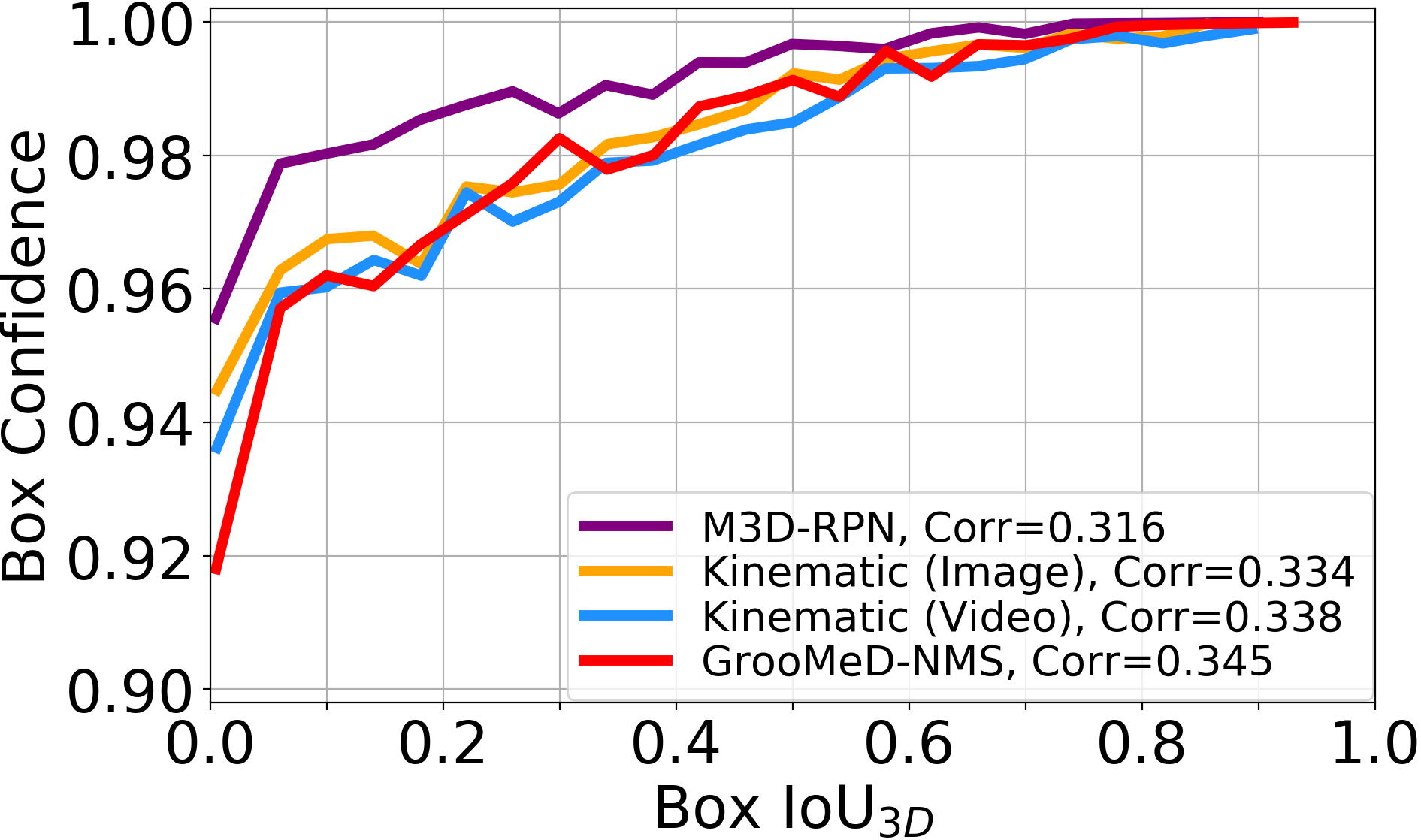}
            \caption{Score-\iouThreeD~plot after the NMS.}
            \label{fig:score_iou}
        \end{figure}

        \noindent\textbf{Score-\iouThreeD~Plot.}
            We further correlate the scores with \iouThreeD~after NMS of our~model with two baselines - M3D-RPN~\cite{brazil2019m3d} and \kinematicImage~\cite{brazil2020kinematic} and also the \kinematicVideo\cite{brazil2020kinematic}~in \myReferFigure{fig:score_iou}. 
            We obtain the best correlation of $0.345$ exceeding the correlations of M3D-RPN, \kinematicImage~and, also \kinematicVideo. This proves that including NMS in the training pipeline is beneficial.
        
        \noindent\textbf{Training and Inference Times.}
            We now compare the training and inference times of including~\methodName~in the pipeline. 
            Warmup training phase takes about $13$ hours to train on a single $12$ GB GeForce GTX Titan-X GPU. 
            Full training phase of \kinematicImage~and~\methodName~takes about $8$ and $8.5$ hours respectively. 
            The inference time per image using \classicalNmsShort~and \methodName~is $0.12$ and $0.15$ ms respectively. 
            \myReferTable{tab:results_kitti_val1_other_nms} suggests that changing the NMS from \methodNameShort~to \classicalNmsShort~during inference does not alter the performance. 
            Then, the inference time of our method is the same as $0.12$ ms.

    \subsection{\kitti~\valTwo~3D~Object Detection}\label{sec:results_kitti_val2}
        \myReferTable{tab:results_kitti_val2} summarizes the results of \threeD~object detection and BEV evaluation on \kitti~\valTwo~Split at two \iouThreeD~thresholds of $0.7$ and $0.5$~\cite{chen2020monopair,brazil2020kinematic}. 
        Again, we use M3D-RPN~\cite{brazil2019m3d} and Kinematic (Image)~\cite{brazil2020kinematic} as our baselines. 
        We evaluate the released model of M3D-RPN~\cite{brazil2019m3d} using the \kitti~metric.~\cite{brazil2020kinematic} does not report \valTwo~results, so we retrain on \valTwo~using their public code. 
        The results in \myReferTable{tab:results_kitti_val2} show that \methodName~performs best in all cases. 
        This is again impressive because the improvements are shown on Moderate and Hard set, consistent with Tabs.~\ref{tab:results_kitti_test} and~\ref{tab:results_kitti_val1}.

    \subsection{Ablation Studies}\label{sec:results_ablation}
        \myReferTable{tab:ablation} compares the modifications of our approach on \kitti~\valOne~Cars. 
        Unless stated otherwise, we stick with the experimental settings described in \myReferSection{sec:experiments}. 
        Using a confidence head (Conf+No NMS) proves beneficial compared to the warmup model (No Conf+No NMS), which is consistent with the observations of \cite{shi2020distance, brazil2020kinematic}. 
        Further, \methodName~on classification scores (denoted by No Conf + NMS) is detrimental as the classification scores are not suited for localization \cite{huang2020epnet, brazil2020kinematic}. 
        Training the warmup model and then finetuning also works better than training without warmup as in~\cite{brazil2020kinematic} since the warmup phase allows \methodName~to carry meaningful grouping of the boxes. 
        
        As described in \myReferSection{sec:pruning}, in addition to Linear, we compare two other functions for pruning function $\prune$: Exponential and Sigmoidal. 
        Both of them do not perform as well as the Linear $\prune$ possibly because they have vanishing gradients close to overlap of zero or one. Grouping and masking both help our model to reach a better minimum. 
        As described in \myReferSection{sec:target_loss}, \imageWise~\ap~loss is better than the Vanilla \ap~loss since it treats boxes of two images differently. 
        \imageWise~\ap~also performs better than the binary cross-entropy (BCE) loss proposed in~\cite{hosang2016convnet, henderson2016end, prokudin2017learning, hosang2017learning}. 
        Using the product of self-balancing confidence and classification scores instead of using them individually as the scores to the NMS in inference is better, consistent with~\cite{tychsen2018improving, shi2020distance, kim2020probabilistic}. 
        Class confidence performs worse since it does not have the localization information while the self-balancing confidence (Pred) gives the localization without considering whether the box belongs to foreground or background.

\section{Conclusions}
    In this paper, we present and integrate \methodName~-- a novel 
    Grouped Mathematically Differentiable NMS for monocular \threeD~object detection, such that the network is trained end-to-end with a loss on the boxes after NMS. 
    We first formulate NMS as a matrix operation and then do unsupervised grouping and masking of the boxes to obtain a simple closed-form expression of the NMS. 
    \methodName~addresses the mismatch between training and inference pipelines and, therefore, forces the network to  select the best \threeD~box in a differentiable manner. 
    As a result, \methodName~achieves state-of-the-art monocular \threeD~object detection results on the \kitti~benchmark dataset. 
    Although our implementation demonstrates monocular \threeD~object detection, \methodName~is fairly generic for other object detection tasks. 
    Future work includes applying this method to tasks such as \lidar-based \threeD~object detection and pedestrian detection.

\clearpage
{\small
\bibliographystyle{ieee_fullname}
\bibliography{references}
}

\clearpage
\appendix 
\twocolumn[\centering \section*{\Large ~\methodName: Grouped Mathematically Differentiable NMS for Monocular 3D Object Detection\\[12pt] Supplementary Material\\[18pt]}]

\renewcommand{\thesection}{A\arabic{section}}

    \begin{table*}[t]
        \caption{Results on using Oracle NMS scores on \apThreeDForty~,\apBevForty~and\apTwoDForty~of \kitti~\valOne~Cars. [Key: \bestKey{Best}]}
        \label{tab:oracle_nms}
        \centering
        \footnotesize
        \setlength\tabcolsep{2.00pt}
        \begin{tabular}{tl m ccc  m ccc m ccct}
            \myTopRule
            \multirow{2}{*}{NMS Scores} & \multicolumn{3}{cm}{\apThreeDForty ($\uparrow$)} & \multicolumn{3}{cm}{\apBevForty ($\uparrow$)} & \multicolumn{3}{ct}{\apTwoDForty ($\uparrow$)}\\ 
            & Easy & Mod & Hard & Easy & Mod & Hard & Easy & Mod & Hard\\ 
            \myTopRule
            \kinematicImage & $18.29$ & $13.55$ & $10.13$ & $25.72$ & $18.82$ & $14.48$ & $93.69$ & $84.07$ & $67.14$ \\
            Oracle \iouTwoD        & $ 9.36$ & $ 9.93$ & $6.40$  & $12.27$ & $10.43$ & $ 8.72$ & $\best{99.18}$ & $\best{95.66}$ & $\best{85.77}$ \\
            Oracle \iouThreeD      & $\best{87.93}$ & $\best{73.10}$ & $\best{60.91}$ & $\best{93.47}$ & $\best{83.61}$ & $\best{71.31}$ & $80.99$ & $78.38$ & $67.66$\\
            \myTopRule
        \end{tabular}
    \end{table*}

\section{Detailed Explanation of NMS as a Matrix Operation}\label{sec:NMS_explanation}
    The rescoring process of the~\classicalNms~is greedy set-based~\cite{prokudin2017learning} and calculates the rescore for a box $i$ (Line $10$ of Alg.~\ref{alg:classical}) as 
    \begin{align}
        r_i &= s_i  \prod\limits_{j\in\vindex_{< i}}\left(1 -\pruneof{\overlap_{ij}}\right),
        \label{eq:classical_all}
    \end{align}
    where $\vindex_{< i}$ is defined as the box indices sampled from $\vindex$ having higher scores than box $i$. For example, let us consider that $\vindex=\{1,5, 7, 9\}$. Then, for $i=7,~ \vindex_{< i} = \{1,5\}$  while for $i=1, \vindex_{< i} = \phi$ with $\phi$ denoting the empty set.
    This is possible since we had sorted the scores $\score$ and $\overlapMat$ in decreasing order (Lines $2$-$3$ of Alg.~\ref{alg:diff_nms}) to remove the non-differentiable hard $\argmax$ operation of the \classicalNms~(Line $6$ of Alg.~\ref{alg:classical}).
    
    \classicalNmsCaps~only takes the overlap with unsuppressed boxes into account. Therefore, we generalize~\eqref{eq:classical_all} by accounting for the 
    effect of all (suppressed and unsuppressed) boxes as
    \begin{align}
        r_i &= s_i  \prod\limits_{j=1}^{i-1}\left(1 -\pruneof{\overlap_{ij}}r_j\right).
        \label{eq:diff_nms_product}
    \end{align}
    The presence of $r_j$ on the RHS of \eqref{eq:diff_nms_product} prevents suppressed boxes $r_j \approx 0$ from influencing other boxes hugely. 
    Let us say we have a box $b_2$ with a high overlap with an unsuppressed box $b_1$. The  \classicalNms~with a threshold pruning function assigns $r_2 = 0$ while \eqref{eq:diff_nms_product} assigns $r_2$ a small non-zero value with a threshold pruning.
    
    Although \eqref{eq:diff_nms_product} keeps $r_i \ge 0$, getting a closed-form recursion in $\rescore$ is not easy because of the product operation. To get a closed-form recursion with addition/subtraction in $\rescore$, we first carry out the polynomial multiplication and then ignore the higher-order terms as
    \begin{align}
        r_i &= s_i \left(1 - \sum\limits_{j=1}^{i-1}\pruneof{\overlap_{ij}}r_j + \mathcal{O}(n^2)\right) \nonumber\\
        &\approx s_i \left(1 - \sum\limits_{j=1}^{i-1}\pruneof{\overlap_{ij}}r_j \right) \nonumber\\
        &\approx s_i -  \sum\limits_{j=1}^{i-1}\pruneof{\overlap_{ij}}r_j.
        \label{eq:diff_nms_rescore_2}
    \end{align}
    Dropping the $s_i$ in the second term of \eqref{eq:diff_nms_rescore_2} helps us get a cleaner form of \eqref{eq:diff_nms_full_again}. Moreover, it does not change the nature of the NMS since the subtraction keeps the relation $r_i \le s_i$ intact as $\pruneof{\overlap_{ij}}$ and $r_j$ are both between $[0, 1]$. 
    
    We can also reach \eqref{eq:diff_nms_rescore_2} directly as follows. \classicalNmsCaps~suppresses a box which has a high \iouTwoD~overlap with \emph{any} of the unsuppressed boxes ($r_j \approx 1$) to zero. 
    We consider \emph{any} as a logical non-differentiable OR operation and use logical OR $\bigvee$ operator's differentiable relaxation as $\sum$~\cite{van2020analyzing, li2019augmenting}. 
    We next use this relaxation with the other expression $\rescore  \le \score$.
    
    When a box shows overlap with more than two unsuppressed boxes, the term $\sum\limits_{j=1}^{i-1}\pruneof{\overlap_{ij}}r_j > 1$ in \eqref{eq:diff_nms_rescore_2} or when a box shows high overlap with one unsuppressed box, the term $s_i < \pruneof{\overlap_{ij}}r_j$. In both of these cases, $r_i < 0$. So, we lower bound \eqref{eq:diff_nms_rescore_2} with a $\max$ operation to ensure that $r_i \ge 0$. Thus,
    \begin{align}
        r_i &\approx \max\left(s_i - \sum\limits_{j=1}^{i-1}\pruneof{\overlap_{ij}}r_j,~0 \right).
        \label{eq:diff_nms_rescore_3}
    \end{align}

    We write the rescores $\rescore$ in a matrix formulation as
    \begin{align}
        \begin{bmatrix}
        r_1 \\
        r_2 \\
        r_3 \\
        \vdots \\
        r_n\\
        \end{bmatrix}
            &\!\approx\! 
        \max\left(\begin{bmatrix}
        c_1\\
        c_2\\
        c_3\\
        \vdots \\
        c_n\\
        \end{bmatrix}
        ,
        \begin{bmatrix}
        0 \\
        0 \\
        0 \\
        \vdots \\
        0\\
        \end{bmatrix}
        \right),
    \end{align}
    with
    \begin{align}
        \begin{bmatrix}
        c_1\\
        c_2\\
        c_3\\
        \vdots \\
        c_n\\
        \end{bmatrix}
        &=
        \begin{bmatrix}
        s_1\\
        s_2\\
        s_3\\
        \vdots \\
        s_n\\
        \end{bmatrix}
        -
        \begin{bmatrix}
        0 & 0 & \dots & 0\\
        \pruneof{\overlap_{21}}\!&\!0\!&\!\dots\!&\!0\\
        \pruneof{\overlap_{31}}\!&\!\pruneof{\overlap_{32}}\!&\!\dots\!&\!0\\
        \vdots\!& \vdots & \vdots\!&\!\vdots \\
        \pruneof{\overlap_{n1}}\!&\!\pruneof{\overlap_{n2}} & \dots & 0 \\
        \end{bmatrix} 
        \begin{bmatrix}
        r_1 \\
        r_2 \\
        r_3 \\
        \vdots \\
        r_n\\
        \end{bmatrix}.
    \end{align}
    
    We next write the above two equations compactly as
    \begin{align}
        \rescore &\approx \max(\score - \pruneMat\rescore,\zeroVector),
        \label{eq:diff_nms_recursive_2}
    \end{align}
    where $\pruneMat$, called the Prune Matrix, is obtained by element-wise operation of the pruning function $\prune$ on $\overlapMatLower$. Maximum operation makes \eqref{eq:diff_nms_recursive_2} non-linear~\cite{kumar2013estimation} and, thus, difficult to solve. 
    
    However, for a differentiable NMS layer, we need to avoid the recursion. Therefore, we first solve \eqref{eq:diff_nms_recursive_2} assuming the $\max$ operation is not present which gives us the solution $\rescore \approx \left( \identity + \pruneMat \right)^{-1}\score$. In general, this solution is not necessarily bounded between $0$ and $1$. Hence, we clip it explicitly to obtain the approximation
    \begin{align}
        \rescore \approx \clip{\left( \identity + \pruneMat \right)^{-1}\score},
        \label{eq:diff_nms_full_again}
    \end{align}
    which we use as the solution to \eqref{eq:diff_nms_recursive_2}.

\section{Loss Functions}\label{sec:appendix_loss}
    We now detail out the loss functions used for training.
    The losses on the boxes before NMS, $\lossBefore$, is given by~\cite{brazil2020kinematic}
    \begin{align}
        \lossBefore &= \loss_\class + \loss_{\twoDMath} + \boxConfidence~\loss_{\threeDMath} \nonumber \\
        &\quad + \lossWeigh_{conf} (1- \boxConfidence),
    \end{align}
    where
    \begin{align}
        \loss_\class &= \text{CE}(b_\class, g_\class), \\
        \loss_{\twoDMath} &= -\log (\text{\iou}(b_{\twoDMath}, g_{\twoDMath})), \\
        \loss_{\threeDMath} &= \text{Smooth-L1} (b_{\threeDMath}, g_{\threeDMath}) \nonumber \\
        &\quad + \lossWeigh_{a} \text{CE}([b_{\theta_a}, b_{\theta_h}], [g_{\theta_a}, g_{\theta_h}]).
    \end{align}
    $\boxConfidence$ is the predicted self-balancing confidence of each box $b$, while $b_{\theta_a}$ and $b_{\theta_h}$ are its orientation bins~\cite{brazil2020kinematic}. 
    $g$ denotes the ground-truth.
    $\lossWeigh_{conf}$ is the rolling mean of most recent $\loss_{\threeDMath}$ losses per mini-batch~\cite{brazil2020kinematic}, while $\lossWeigh_{a}$ denotes the weight of the orientation bins loss.
    CE and Smoooth-L1 denote the Cross Entropy and Smooth L1 loss respectively.
    Note that we apply \twoD~and \threeD~regression losses as well as the confidence losses only on the foreground boxes.
    
    As explained in \myReferSection{sec:target_loss}, the loss on the boxes after NMS, $\lossAfter$, is the \imageWise~\aploss, which is given by
    \begin{align}
        \lossAfter&=\!\loss_{\imageWise}\!=\!\frac{1}{N}\sum_{m=1}^N\!\apMath(\rescore^{(m)}, \text{target}(\boxes^{(m)})),
    \end{align}
    
    Let $\lossWeigh$ be the weight of the $\lossAfter$ term.
    Then, our overall loss function is given by 
    \begin{align}
        \loss &= \lossBefore + \lossWeigh \lossAfter \\
              &= \loss_\class + \loss_{\twoDMath} + \boxConfidence~\loss_{\threeDMath} + \lossWeigh_{conf} (1- \boxConfidence) \nonumber \\
              &\quad + \lossWeigh \loss_{\imageWise} \\
              &= \text{CE}(b_\class, g_\class) -  \log (\text{\iou}(b_{\twoDMath}, g_{\twoDMath})) \nonumber \\
              &\quad + \boxConfidence~\text{Smooth-L1} (b_{\threeDMath}, g_{\threeDMath}) \nonumber \\
              &\quad + \lossWeigh_{a}~\boxConfidence~\text{CE}([b_{\theta_a}, b_{\theta_h}], [g_{\theta_a}, g_{\theta_h}]) \nonumber \\
              &\quad + \lossWeigh_{conf} (1- \boxConfidence) + \lossWeigh \loss_{\imageWise}.
    \end{align}
    We keep $\lossWeigh_{a}= 0.35$ following~\cite{brazil2020kinematic}
    and $\lossWeigh = 0.05$. 
    Clearly, all our losses and their weights are identical to~\cite{brazil2020kinematic} except $\loss_{\imageWise}$.

    \begin{table*}[t]
        \caption{\apThreeDForty~and \apBevForty~comparisons with other NMS during inference on \kitti~\valOne~Cars.}
        \label{tab:results_kitti_val1_other_nms_detailed}
        \centering
        \footnotesize
        \setlength\tabcolsep{0.1cm}
        \begin{tabular}{tl t c m ccc t ccc m ccc t ccct}
            \myTopRule
            \addlinespace[0.01cm]
            \multirow{3}{*}{} &  \multirow{3}{*}{\shortstack{Inference\\NMS}} & \multicolumn{6}{cm}{\iouThreeD~$\geq 0.7$} & \multicolumn{6}{ct}{\iouThreeD~$\geq 0.5$}\\\cline{3-14}
            & & \multicolumn{3}{ct}{\apThreeDForty ($\uparrow$)} & \multicolumn{3}{cm}{\apBevForty ($\uparrow$)} & \multicolumn{3}{ct}{\apThreeDForty ($\uparrow$)} & \multicolumn{3}{ct}{\apBevForty ($\uparrow$)}\\
            & & Easy & Mod & Hard & Easy & Mod & Hard & Easy & Mod & Hard & Easy & Mod & Hard\\ 
            \myTopRule
            \kinematicImage~\cite{brazil2020kinematic}      &\classicalNmsShortCaps& $18.28$        & $13.55$        & $10.13$       & $25.72$        & $18.82$       & $14.48$       & $54.70$        & $39.33$        & $31.25$        & $60.87$        & $44.36$       & $34.48$       \\
            \kinematicImage~\cite{brazil2020kinematic}      &\softNmsShortCaps~\cite{bodla2017soft}                 & $18.29$        & $13.55$        & $10.13$       & $25.71$        & $18.81$        & $14.48$       & $54.70$       & $39.33$        & $31.26$        & $60.87$        & $44.36$       & $34.48$       \\
            \kinematicImage~\cite{brazil2020kinematic}      &\distanceNmsShortCaps~\cite{shi2020distance}          & $18.25$        & $13.53$        & $10.11$       & $25.71$        & $18.82$       & $14.48$        & $54.70$       & $39.33$        & $31.26$        & $60.87$        & $44.36$       & $34.48$       \\
            \kinematicImage~\cite{brazil2020kinematic}      &\methodNameShort~     & $18.26$        & $13.51$        & $10.10$       & $25.67$        & $18.77$       & $14.44$        & $54.59$       & $39.25$        & $31.18$        & $60.78$        & $44.28$       & $34.40$       \\
            \hline
            \methodName~                     &\classicalNmsShortCaps& $19.67$ & $14.31$  & $11.27$ & $27.38$ & $19.75$  & $15.93$ & $55.64$ & $41.08$ & $32.91$& $61.85$ & $44.98$ & $36.31$\\
            \methodName~                     &\softNmsShortCaps~\cite{bodla2017soft}& $19.67$ & $14.31$  & $11.27$ & $27.38$ & $19.75$  & $15.93$ & $55.64$ & $41.08$ & $32.91$& $61.85$ & $44.98$ & $36.31$\\
            \methodName~                     &\distanceNmsShortCaps~\cite{shi2020distance}          & $19.67$ & $14.31$  & $11.27$ & $27.38$ & $19.75$  & $15.93$ & $55.64$ & $41.08$ & $32.91$& $61.85$ & $44.98$ & $36.31$\\
            \methodName~                     &\methodNameShort~     & $19.67$ & $14.32$  & $11.27$ & $27.38$ & $19.75$  & $15.92$ & $55.62$ & $41.07$ & $32.89$& $61.83$ & $44.98$ & $36.29$\\
            \myTopRule
        \end{tabular}
    \end{table*}

\section{Additional Experiments and Results}\label{sec:additional_exp}
    We now provide additional details and results evaluating our system's performance.

    \subsection{Training}\label{sec:training_additional}
        Training images are augmented using random flipping with probability $0.5$~\cite{brazil2020kinematic}.
        Adam optimizer~\cite{kingma2014adam} is used with batch size $2$, weight-decay $5\times10^{-4}$ and gradient clipping of $1$~\cite{brazil2019m3d, brazil2020kinematic}.
        Warmup starts with a learning rate $4 \times 10^{-3}$ following a poly learning policy with power $0.9$~\cite{brazil2020kinematic}.
        Warmup and full training phases take $80k$ and $50k$ mini-batches respectively for \valOne~and~\valTwo~Splits~\cite{brazil2020kinematic} while take $160k$ and $100k$ mini-batches for Test Split. 

    \subsection{\kitti~\valOne~Oracle NMS Experiments}\label{sec:results_oracle_additional}
        As discussed in \myReferSection{sec:Introduction}, to understand the effects of an inference-only NMS on \twoD~and \threeD~object detection, we conduct a series of oracle experiments.
        We create an oracle NMS by taking the Val Car boxes of \kitti~\valOne~Split from the baseline \kinematicImage~model \textit{before} NMS and replace their scores with their true \iouTwoD~or \iouThreeD~with the ground-truth, respectively. 
        Note that this corresponds to the oracle because we do not know the ground-truth boxes during inference.
        We then pass the boxes with the oracle scores through the \classicalNms~and report the results in \myReferTable{tab:oracle_nms}.
        
        The results show that the \apthreeD~increases by a staggering $>60$ \ap~on Mod cars when we use oracle \iouThreeD~as the NMS score.
        On the other hand, we only see an increase in \apTwoD~by $\approx 11$ \ap~on Mod cars when we use oracle \iouTwoD~as the NMS score.
        Thus, the relative effect of using oracle \iouThreeD~NMS scores on \threeD~detection is more significant than using oracle \iouTwoD~NMS scores on \twoD~detection.
        In other words, the mismatch is greater between classification and \threeD~localization compared to the mismatch between classification and \twoD~localization.

    \subsection{\kitti~\valOne~3D~Object Detection}\label{sec:results_kitti_val1_additional}
    
    \textbf{Comparisons with other NMS.}
        We compare our method with the other NMS---\classicalNmsShort, \softNmsShortCaps~\cite{bodla2017soft} and \distanceNmsCaps~\cite{shi2020distance} and report the detailed results in \myReferTable{tab:results_kitti_val1_other_nms_detailed}. 
        We use the publicly released \softNmsCaps~code and \distanceNmsCaps~code from the respective authors.
        The \distanceNmsCaps~model uses the class confidence scores divided by the uncertainty in $z$ (the most erroneous dimension in \threeD~localization~\cite{simonelli2020demystifying}) of a box as the \distanceNmsCaps~\cite{shi2020distance} input. 
        Our model does not predict the uncertainty in $z$ of a box
        but predicts its self-balancing confidence (the \threeD~localization score). 
        Therefore, we use the class confidence scores multiplied by the self-balancing confidence as the \distanceNmsCaps~input.
        
        The results in \myReferTable{tab:results_kitti_val1_other_nms_detailed} show that NMS inclusion in the training pipeline benefits the performance, unlike~\cite{bodla2017soft}, which suggests otherwise.
        Training with \methodName~helps because the network gets an additional signal through the \methodName~layer whenever the best-localized box corresponding to an object is not selected. 
        Moreover, \myReferTable{tab:results_kitti_val1_other_nms_detailed} suggests that we can replace \methodName~with the~\classicalNms~in inference as the performance is almost the same even at \iouThreeD$=0.5$.

    \textbf{How good is the \classicalNms~approximation?}
        \methodName~uses several approximations to arrive at the matrix solution~\eqref{eq:diff_nms_full_again}.
        We now compare how good these approximations are with the \classicalNms. 
        Interestingly, \myReferTable{tab:results_kitti_val1_other_nms_detailed} shows that \methodName~is an excellent approximation to the \classicalNms~as the performance does not degrade after changing the NMS in inference.

    \subsection{\kitti~\valOne~Sensitivity Analysis}
        There are a few adjustable parameters for the \methodName, such as the NMS threshold $\nmsThresh$, valid box threshold $\validBoxThresh$, the maximum group size $\alpha$, the weight $\lossWeigh$ for the $\lossAfter$, and $\beta$. We carry out a sensitivity analysis to understand how these parameters affect performance and speed, and how sensitive the algorithm is to these parameters.
        
        \textbf{Sensitivity to NMS Threshold.} 
        We show the sensitivity to NMS threshold $\nmsThresh$ in  \myReferTable{tab:sensitivity_to_nms_thresh}.
        The results in \myReferTable{tab:sensitivity_to_nms_thresh} show that the optimal $\nmsThresh= 0.4$.
        This is also the $\nmsThresh$ in~\cite{brazil2019m3d, brazil2020kinematic}.
        
        \textbf{Sensitivity to Valid Box Threshold.} 
        We next show the sensitivity to valid box threshold $\validBoxThresh$ in \myReferTable{tab:sensitivity_to_v}. 
        Our choice of $\validBoxThresh= 0.3$ performs close to the optimal choice.

        \begin{table}[!t]
            \caption{\apThreeDForty~and \apBevForty~ variation with $\nmsThresh$ on \kitti~\valOne~Cars. [Key: \bestKey{Best}]}
            \label{tab:sensitivity_to_nms_thresh}
            \centering
            \footnotesize
            \setlength\tabcolsep{2.00pt}
            \begin{tabular}{tl m ccc  m ccct}
                \myTopRule
                \multirow{2}{*}{ } & \multicolumn{3}{cm}{\apThreeDForty ($\uparrow$)} & \multicolumn{3}{ct}{\apBevForty ($\uparrow$)}\\ 
                & Easy & Mod & Hard & Easy & Mod & Hard\\ 
                \myTopRule
                $\nmsThresh=0.3$ & $17.49$ & $13.32$ & $10.54$ & $26.07$ & $18.94$ & $14.61$\\
                $\nmsThresh=\best{0.4}$ & $\best{19.67}$ & $\best{14.32}$ & $\best{11.27}$ & $\best{27.38}$ & $\best{19.75}$ & $\best{15.92}$\\
                $\nmsThresh=0.5$ & $19.65$ & $13.93$ & $11.09$ & $26.15$ & $19.15$ & $14.71$\\
                \myTopRule
            \end{tabular}
        \end{table}
    
        \begin{table}[!t]
            \caption{\apThreeDForty~and \apBevForty~variation with $\validBoxThresh$ on \kitti~\valOne~Cars. [Key: \bestKey{Best}]}
            \label{tab:sensitivity_to_v}
            \centering
            \footnotesize
            \setlength\tabcolsep{2.00pt}
            \begin{tabular}{tl m ccc  m ccct}
                \myTopRule
                \multirow{2}{*}{ } & \multicolumn{3}{cm}{\apThreeDForty ($\uparrow$)} & \multicolumn{3}{ct}{\apBevForty ($\uparrow$)}\\ 
                & Easy & Mod & Hard & Easy & Mod & Hard\\ 
                \myTopRule
                $\validBoxThresh= 0.01$& $13.71$ & $9.65$ & $7.24$ & $17.73$ & $12.47$ & $9.36$\\
                $\validBoxThresh= 0.1$ & $19.37$ & $13.99$ & $10.92$ & $26.95$ & $19.84$ & $15.40$\\
                $\validBoxThresh= 0.2$ & $19.65$ & $14.31$ & $11.24$ & $27.35$ & $19.73$ & $15.89$\\
                $\validBoxThresh= 0.3$ & $19.67$ & $14.32$ & $11.27$ & $27.38$ & $19.75$ & $15.92$\\
                $\validBoxThresh= 0.4$ & ${19.67}$ & ${14.33}$ & ${11.28}$ & ${27.38}$ & ${19.76}$ & ${15.93}$\\
                $\validBoxThresh= 0.5$ & $19.67$ & $14.33$ & $11.28$ & $27.38$ & $19.76$ & $15.93$\\
                $\validBoxThresh= \best{0.6}$ & $\best{19.67}$ & $\best{14.33}$ & $\best{11.29}$ & $\best{27.39}$ & $\best{19.77}$ & $\best{15.95}$\\
                \myTopRule
            \end{tabular}
        \end{table}

\begin{figure} [t]
            \centering
            \includegraphics[width=0.7\linewidth]{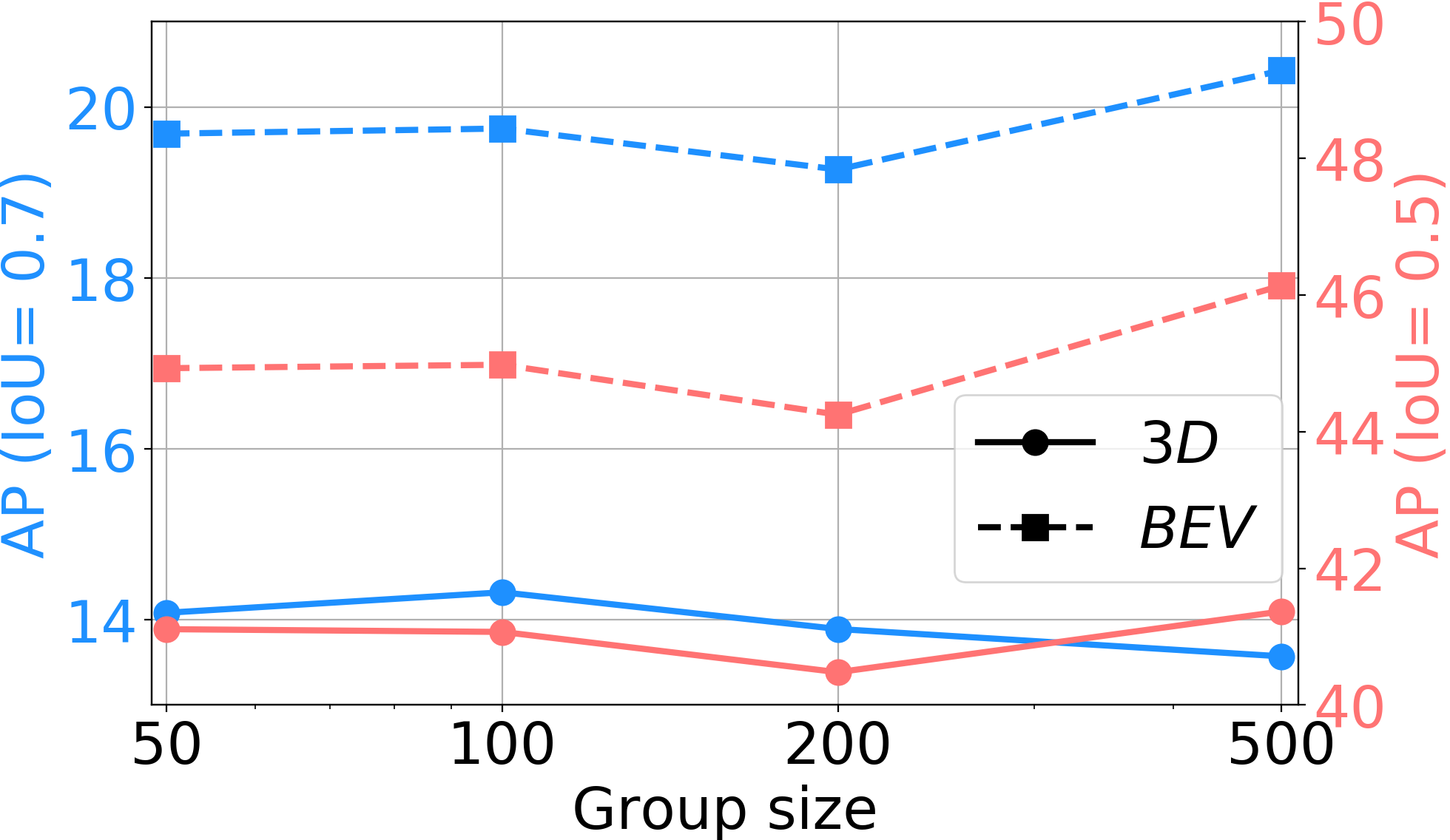}
            \caption{\apThreeDForty~and~\apBevForty~Variation with $\groupSize$ on Moderate Cars of \kitti~\valOne~Split.}
            \label{fig:sensitivity_to_group_size}
        \end{figure}

        \textbf{Sensitivity to Maximum Group Size.}
            Grouping has a parameter group size $(\groupSize)$. 
            We vary this parameter and report \apThreeDForty~and~\apBevForty~at two different \iouThreeD~thresholds on Moderate Cars of \kitti~\valOne~Split in \myReferFigure{fig:sensitivity_to_group_size}. 
            We note that the best \apThreeDForty~performance is obtained at $\groupSize= 100 $ and we, therefore, set $\groupSize= 100$ in our experiments.
        
        \textbf{Sensitivity to Loss Weight.}
            We now show the sensitivity to loss weight $\lossWeigh$ in \myReferTable{tab:sensitivity_to_lambda}. 
            Our choice of $\lossWeigh= 0.05$ is the optimal value.
        
        \textbf{Sensitivity to Best Box Threshold.}
            We now show the sensitivity to the best box threshold $\beta$ in \myReferTable{tab:sensitivity_to_beta}. 
            Our choice of $\beta= 0.3$ is the optimal value.
        
        \textbf{Conclusion.}
            Our method has minor sensitivity to $\nmsThresh, \groupSize, \lossWeigh$ and $\beta$, which is common in object detection. 
            Our method is not as sensitive to $\validBoxThresh$ since it only decides a box's validity. 
            Our parameter choice is either at or close to the optimal.
            The inference speed is only affected by $\groupSize$. Other parameters are used in training or do not affect inference speed.
    
        \begin{table}[!t]
            \caption{\apThreeDForty~and \apBevForty~variation with $\lossWeigh$ on \kitti~\valOne~Cars. [Key: \bestKey{Best}]}
            \label{tab:sensitivity_to_lambda}
            \centering
            \footnotesize
            \setlength\tabcolsep{2.00pt}
            \begin{tabular}{tl m ccc  m ccct}
                \myTopRule
                \multirow{2}{*}{ } & \multicolumn{3}{cm}{\apThreeDForty ($\uparrow$)} & \multicolumn{3}{ct}{\apBevForty ($\uparrow$)}\\ 
                & Easy & Mod & Hard & Easy & Mod & Hard\\ 
                \myTopRule
                $\lossWeigh= 0$ & $19.16$ & $13.89$ & $10.96$ & $27.01$ & $19.33$ & $14.84$\\
                $\best{\lossWeigh= 0.05}$ & $\best{19.67}$ & $\best{14.32}$ & $\best{11.27}$ & $\best{27.38}$ & $\best{19.75}$ & $\best{15.92}$\\
                $\lossWeigh= 0.1$ & $17.74$ & $13.61$ & $10.81$ & $25.86$	& $19.18$ & $15.57$\\
                $\lossWeigh= 1$   & $10.08$ & $ 7.26$ & $ 6.00$ & $14.44$ & $10.55$ & $8.41$\\
                \myTopRule
            \end{tabular}
        \end{table}

        \begin{table}[!t]
            \caption{\apThreeDForty~and \apBevForty~variation with $\beta$ on \kitti~\valOne~Cars. [Key: \bestKey{Best}]}
            \label{tab:sensitivity_to_beta}
            \centering
            \footnotesize
            \setlength\tabcolsep{2.00pt}
            \begin{tabular}{tl m ccc  m ccct}
                \myTopRule
                \multirow{2}{*}{ } & \multicolumn{3}{cm}{\apThreeDForty ($\uparrow$)} & \multicolumn{3}{ct}{\apBevForty ($\uparrow$)}\\ 
                & Easy & Mod & Hard & Easy & Mod & Hard\\ 
                \myTopRule
                $\beta= 0.1$ & $18.09$ & $13.64$ & $10.21$ & $26.52$ & $19.50$ & $15.74$\\ 
                $\best{\beta= 0.3}$ & $\best{19.67}$ & $\best{14.32}$ & $\best{11.27}$ & $\best{27.38}$ & $\best{19.75}$ & $\best{15.92}$\\ 
                $\beta= 0.4$ & $18.91$ & $14.02$ & $11.15$ & $27.11$ & $19.64$ & $15.90$\\
                $\beta= 0.5$ & $18.49$ & $13.66$ & $10.96$ & $27.01$ & $19.47$ & $15.79$\\
                \myTopRule
            \end{tabular}
        \end{table}

    \subsection{Qualitative Results}
\begin{figure*}[!tb]
            \centering
            \begin{subfigure}{\linewidth}
              \includegraphics[width=\linewidth]{images/qualitative/000514.png}
            \end{subfigure}
            \begin{subfigure}{\linewidth}
              \includegraphics[width=\linewidth]{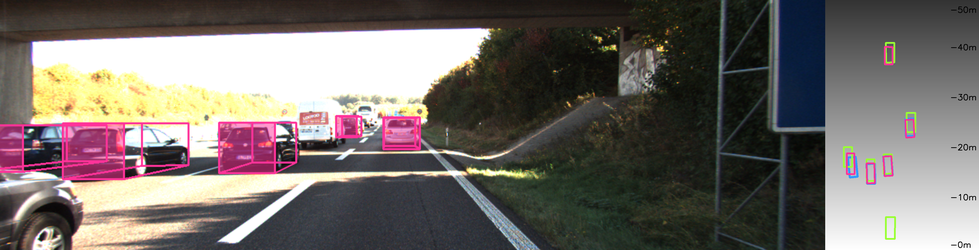}
            \end{subfigure}
            \begin{subfigure}{\linewidth}
              \includegraphics[width=\linewidth]{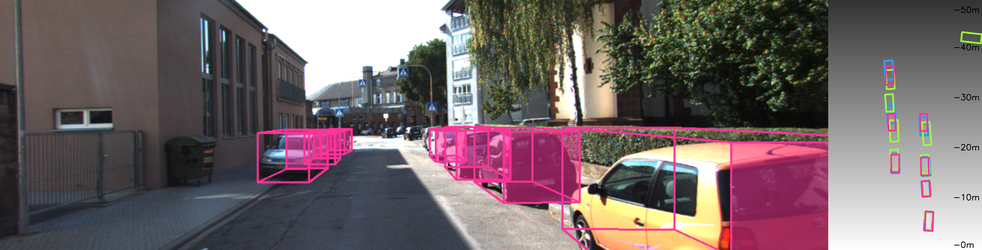}
            \end{subfigure}
            \begin{subfigure}{\linewidth}
              \includegraphics[width=\linewidth]{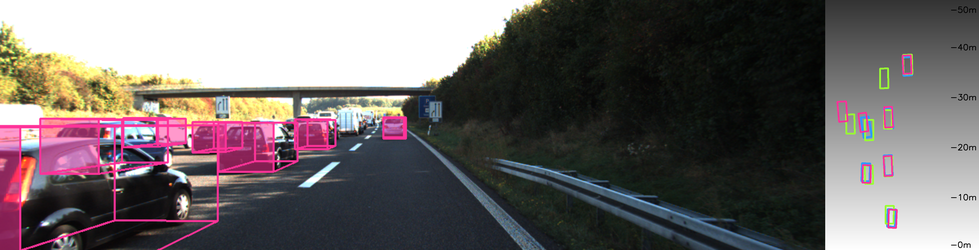}\\
            \end{subfigure}
            \caption{\textbf{Qualitative Results} (Best viewed in color). We depict the predictions of \textcolor{my_orange_2}{\methodName} in image view on the left and the predictions of \textcolor{my_orange_2}{\methodName}, \textcolor{my_blue}{\kinematicImage}~\cite{brazil2020kinematic}, and \textcolor{forward_color}{Ground Truth} in BEV on the right. In general, \textcolor{my_orange_2}{\methodName} predictions are more closer to the \textcolor{forward_color}{ground truth} than \textcolor{my_blue}{\kinematicImage}~\cite{brazil2020kinematic}.}
            \label{fig:qualitative}
        \end{figure*}

        We next show some qualitative results of models trained on \kitti~\valOne~Split in \myReferFigure{fig:qualitative}. We depict the predictions of \methodName~in image view on the left and the predictions of \methodName, \kinematicImage~\cite{brazil2020kinematic}, and ground truth in BEV on the right. In general, \methodName~predictions are more closer to the ground truth than \kinematicImage~\cite{brazil2020kinematic}.

    \subsection{Demo Video of \methodName}   
        We next include a short demo video of our \methodName~model trained on \kitti~\valOne~Split. 
        We run our trained model independently on each frame of the three \kitti~raw~\cite{geiger2013vision} sequences - \textsc{2011\_10\_03\_drive\_0047}, \textsc{2011\_09\_29\_drive\_0026} and \textsc{2011\_09\_26\_drive\_0009}.
        None of the frames from these three raw sequences appear in the training set of \kitti~\valOne~Split.
        We use the camera matrices available with the raw sequences but do not use any temporal information. 
        Overlaid on each frame of the raw input videos, we plot the projected \threeD~boxes of the predictions and also plot these \threeD~boxes in the BEV. 
        We set the frame rate of this demo at $10$ fps.
        The demo is also available in HD at \url{https://www.youtube.com/watch?v=PWctKkyWrno}.
        In the demo video, notice that the orientation of the boxes are stable despite not using any temporal information.

\section*{Acknowledgements}

    This research was partially sponsored by Ford Motor Company and the Army Research Office under Grant Number W911NF-18-1-0330. 
    This document's views and conclusions are those of the authors and do not represent the official policies, either expressed or implied, of the Army Research Office or the U.S.~Government.
    
    We thank Mathieu Blondel and Quentin Berthet from Google Brain, Paris, for several useful discussions on differentiable ranking and sorting.
    We also discussed the logical operators' relaxation with Ashim Gupta from the University of Utah.
    Armin Parchami from Ford Motor Company suggested the learnable NMS paper~\cite{hosang2017learning}.
    Enrique Corona and Marcos Paul Gerardo Castro from Ford Motor Company provided feedback during the development of this work.
    Shengjie Zhu from the Computer Vision Lab at Michigan State University proof-read our manuscript and suggested several changes.
    We also thank Xuepeng Shi from University College London for sharing the \distanceNmsCaps~\cite{shi2020distance}~code for bench-marking.
    We finally acknowledge anonymous reviewers for their feedback that helped in shaping the final manuscript.
    
\end{document}